\documentclass{article}

    \PassOptionsToPackage{numbers, sort&compress}{natbib}

\usepackage[eandd, preprint]{neurips_2026}

\usepackage[utf8]{inputenc} 
\usepackage[T1]{fontenc}    
\usepackage{hyperref}       
\usepackage{url}            
\usepackage{booktabs}       
\usepackage{amsfonts}       
\usepackage{nicefrac}       
\usepackage{microtype}      
\usepackage{xcolor}         
\usepackage{wrapfig}

\usepackage{graphicx}
\usepackage{xspace}
\usepackage{booktabs}
\usepackage{multirow}
\usepackage{enumitem}
\usepackage{xcolor}

\usepackage{microtype}
\usepackage{inconsolata}
\usepackage{makecell}

\usepackage[table]{xcolor}
\definecolor{lightestgray}{gray}{0.92}

\usepackage{placeins}
\usepackage{graphicx}
\usepackage{float}
\usepackage{xurl}
\usepackage{caption}

\usepackage{xcolor}

\definecolor{ansgreen}{RGB}{0,140,0}
\definecolor{evidenceRed}{RGB}{200,40,40}

\usepackage[most]{tcolorbox}

\tcbset{
qaexample/.style={
colback=gray!2,
colframe=black,
boxrule=0.6pt,
arc=2pt,
left=6pt,
right=6pt,
top=6pt,
bottom=6pt,
}
}
\newcommand{\diagrams}{\textsc{DIAGRAMS}\xspace}
\newcommand{\benchmark}{\textsc{DRAGON}\xspace}

\title{\benchmark : A Benchmark for Evidence-Grounded Visual Reasoning over Diagrams}

\newcommand{\logosup}[1]{\raisebox{0.6ex}{\hbox{\scriptsize #1}}}
\newcommand{\asulogo}{\includegraphics[height=8pt]{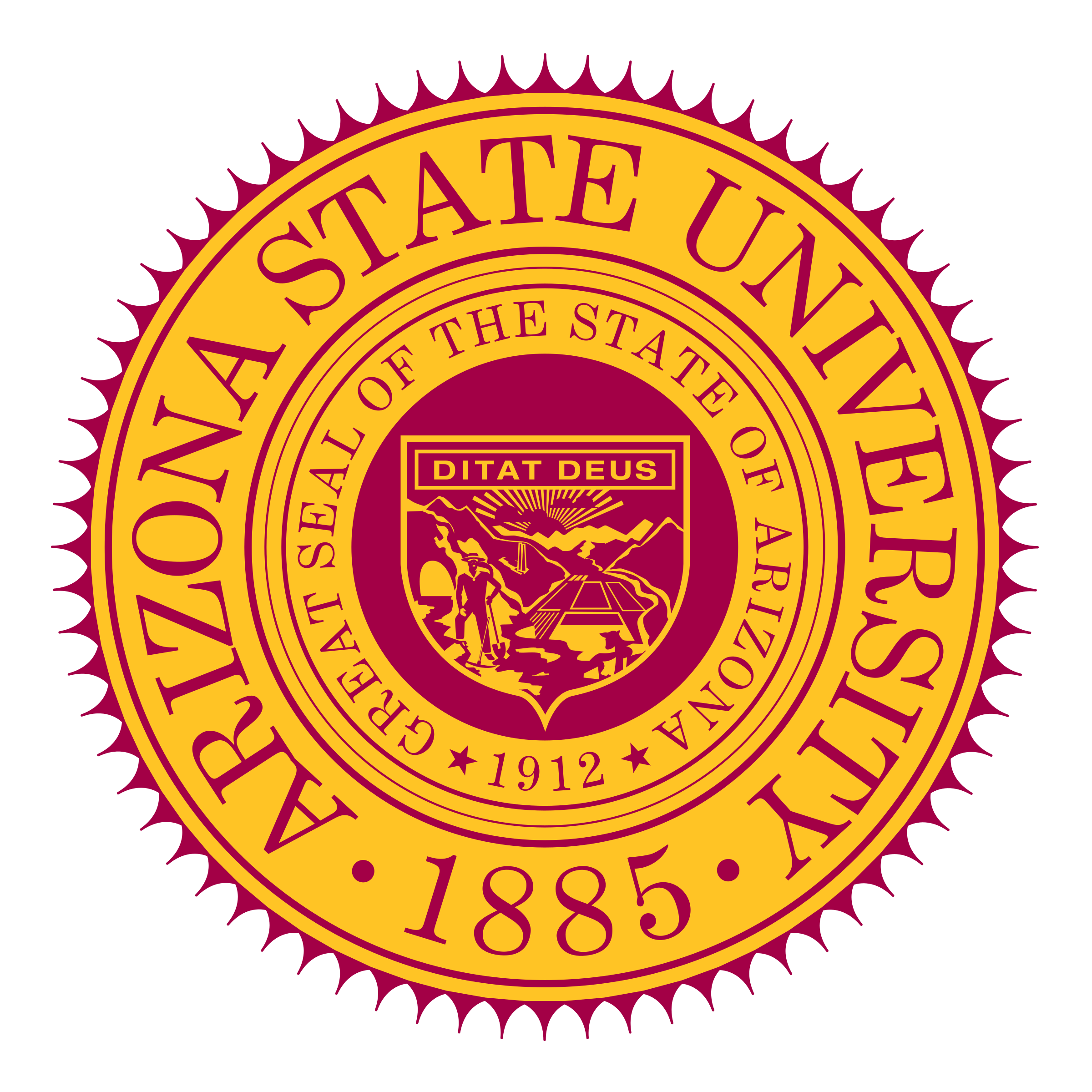}}

\newcommand{\adobelogo}{\includegraphics[height=8pt]{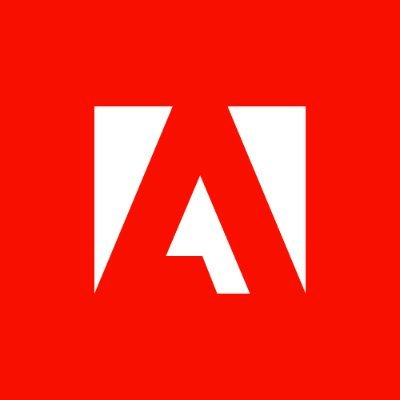}}
\newcommand{\umclogo}{\includegraphics[height=8pt]{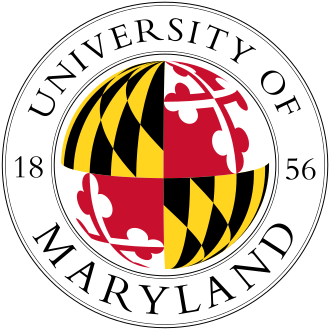}}

\newcommand{\asu}{\logosup{\asulogo}}
\newcommand{\adobe}{\logosup{\adobelogo}}
\newcommand{\umc}{\logosup{\umclogo}}

\author{
\textbf{Anirudh Iyengar Kaniyar Narayana Iyengar}\asu\textsuperscript{*},
\textbf{Tampu Ravi Kumar}\asu\textsuperscript{*},
\textbf{Gaurav Najpande}\asu,\\
\textbf{Manan Suri}\umc,
\textbf{Dinesh Manocha}\umc,
\textbf{Puneet Mathur}\adobe,
\textbf{Vivek Gupta}\asu
\\[2pt]
\asu~Arizona State University 
\adobe~Adobe Research 
\umc~University of Maryland
\\[0pt]
\texttt{\footnotesize akaniyar@asu.edu, traviku2@asu.edu, gnajpand@asu.edu,
manans@umd.edu,}\\[-3pt]
\texttt{\footnotesize
dmanocha@umd.edu,
puneetm@adobe.com, vgupt140@asu.edu
}
}
\begin{document}
\maketitle
{\renewcommand{\thefootnote}{}%
 \footnotetext{$^{*}$These authors contributed equally to this work.}}


\begin{center}
\includegraphics[width=1\linewidth]{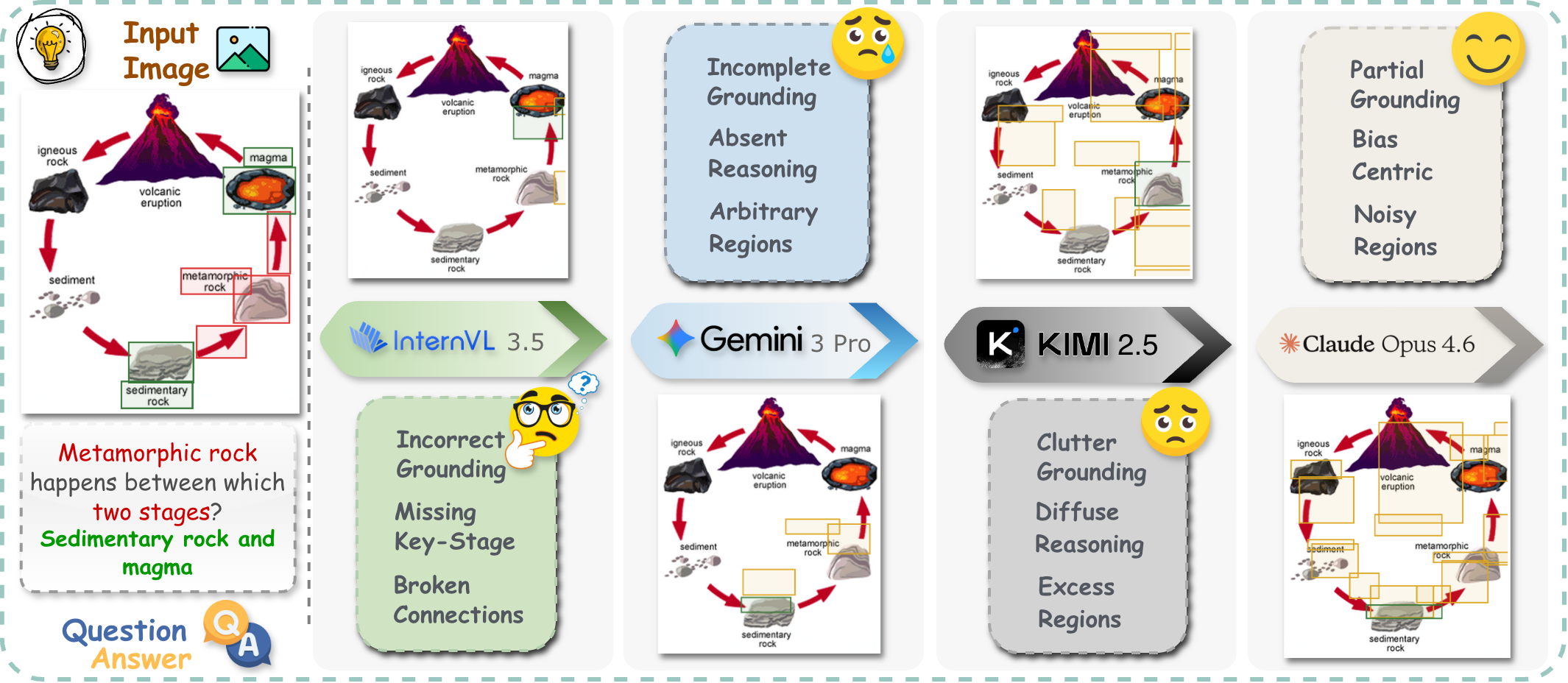}
\vspace{-0.5em}
\captionof{figure}{{Examples of leading models on the Evidence-Grounded Visual Reasoning (\benchmark) benchmark.} \benchmark contains diverse diagrams that pose challenges to current models.}
\end{center}

\begin{abstract}
Diagram question answering (DQA) requires models to interpret structured visual 
representations such as charts, maps, infographics, circuit schematics, and scientific 
diagrams. Recent vision-language models (VLMs) often achieve high answer accuracy on 
these tasks, yet correct answers do not guarantee that models ground their reasoning 
in the diagram regions that support the prediction. Models may instead rely on textual 
correlations or dataset artifacts without identifying the visual evidence required to 
verify the answer. This limitation prevents reliable evaluation of diagram reasoning 
and reduces interpretability. We introduce \textbf{\benchmark}, a benchmark for 
evaluating evidence-grounded visual reasoning in diagrams. Given a diagram, a 
question, and the correct answer, a model must predict bounding boxes that correspond 
to the visual elements required to justify the answer. These evidence regions may 
include answer-bearing components, textual labels, legends, axes, connectors, and 
other supporting structures involved in the reasoning process. The \benchmark dataset 
contains 11,664 annotated question instances from six diagram QA datasets: ChartQA, 
Circuit-VQA, InfographicsVQA, MapIQ, MapWise, and AI2D, with a 2,445-instance test 
set carrying human-verified evidence annotations and a standardized evaluation 
framework. Our evaluation on recent VLMs reveals that even
Claude Opus 4.6 achieves a Grounding IoU of only 23.8\% on AI2D, showing that 
faithful visual grounding remains an open challenge. \benchmark supports future 
research on models that ground their predictions in visual evidence.

\end{abstract}


\section{Introduction}

Diagrams such as charts, maps, infographics, circuit schematics, and scientific illustrations are fundamental to communicating structured information. Interpreting these artifacts requires integrating textual cues, spatial layouts, and structural relationships among visual elements. Diagram question answering (DQA) tasks evaluate a model’s ability to reason over such structured visual representations and have been studied across datasets such as AI2D and ChartQA \cite{kembhavi2016diagram, masry2022chartqa}. Recent vision-language models achieve strong answer accuracy on these benchmarks, yet correct predictions do not necessarily indicate that a model grounds its reasoning in the visual evidence present in the diagram. Human reasoning over diagrams requires explicit visual verification. A person answering a diagram-based question first identifies relevant regions, interprets labels and structural elements, and then combines multiple visual cues to reach a conclusion. Models, however, may produce plausible answers by exploiting textual correlations or dataset biases without identifying the diagram regions that support the answer. Prior work has shown that models can generate correct responses without localizing the visual evidence underlying their predictions \cite{park2018multimodal, hudson2019gqa}. As a result, existing diagram QA evaluations provide limited insight into whether models perform faithful visual reasoning.

To address this gap, we introduce the task of \emph{evidence-grounded reasoning in diagrams}. In this setting, the input consists of a diagram, a question, and the correct answer. The model must identify the visual evidence required to verify the answer by predicting a set of bounding boxes corresponding to the diagram elements involved in the reasoning process, including answer regions, textual labels, legends, axes, connectors, and other supporting structures. This formulation enables direct evaluation of reasoning-level visual grounding rather than answer-only localization.

We present \benchmark, a benchmark for evaluating evidence-grounded reasoning in diagrams. We construct the dataset by curating instances from six established diagram QA datasets: AI2D \cite{kembhavi2016diagram}, ChartQA \cite{masry2022chartqa}, Circuit-VQA \cite{mehta2024circuitvqa}, InfographicsVQA \cite{mathew2022infographicvqa}, MapIQ \cite{srivastava2025mapiq}, and MapWise \cite{mukhopadhyay2025mapwise}. Starting from publicly available question–answer pairs, we annotate the minimal set of diagram regions required to verify each answer. The resulting dataset contains 11,664 annotated question instances spanning multiple diagram domains. We release a benchmark test set of 2,445 human-verified instances together with a standardized evaluation framework for reasoning-level diagram grounding. \noindent{Our main contributions are:}

\begin{enumerate}[leftmargin=*]
\vspace{-0.3em}
\item 
We introduce \emph{evidence-grounded reasoning in diagrams}, a new evaluation task that requires models to localize the visual regions that justify answers to diagram questions instead of relying only on answer prediction.
\vspace{-0.3em}
\item 
We introduce \benchmark, a human-verified multi-domain reasoning benchmark that contains a test set of 2,445 instances with annotated visual evidence regions.
\vspace{-0.3em}
\item 
We design an evaluation framework that studies evidence localization using three prompting strategies: EDGE, SAGE, and VERGE. Our analysis reveals consistent grounding failures even when models produce correct answers.

\end{enumerate}

\section{Related Work}

Research on grounding language in visual regions established the foundation for 
linking textual descriptions to spatial evidence in images. ReferItGame formalized 
referring expression comprehension at scale \cite{kazemzadeh2014referitgame}, and 
subsequent work improved contextual reasoning for referring expressions 
\cite{yu2016modeling}. Methods such as Grad-CAM \cite{selvaraju2017grad} and RISE 
\cite{petsiuk2018rise} highlight image regions that influence model outputs, while 
multimodal explanation frameworks pair textual justifications with visual evidence 
\cite{park2018multimodal}. Visual reasoning benchmarks further extended multimodal 
evaluation: GQA evaluates compositional reasoning over objects and relations 
\cite{hudson2019gqa}, VCR studies visual commonsense reasoning through rationale 
selection \cite{zellers2019recognition}, and A-OKVQA investigates reasoning that 
combines visual perception with external knowledge \cite{schwenk2022okvqa}. Large vision-language models such as LLaVA \cite{liu2023visual} and BLIP-2 
\cite{li2023blip} advance multimodal reasoning through visual instruction tuning 
and language-image pretraining, and models 
such as Qwen-VL introduce strong localization capabilities in multimodal 
architectures \cite{bai2023qwenvlversatilevisionlanguagemodel}. More recently, GRIT trains models to interleave natural language with bounding box coordinates within reasoning chains, enabling grounded reasoning without explicit annotation 
\cite{fan2025grit}. These developments demonstrate that spatial grounding supports 
multimodal reasoning, but existing evaluations rarely test whether predicted regions correspond to the evidence required to verify an answer.

Recent work has pushed visual grounding further by coupling it with reinforcement 
learning and iterative reasoning. ViGoRL trains vision-language models with RL to 
anchor each reasoning step to specific visual coordinates \cite{sarch2025grounded}. 
VGR detects relevant regions during thinking and replays visual memory from critical areas to assist reasoning \cite{wang2025vgr}. RegionReasoner enforces grounded reasoning by requiring each reasoning trace 
to cite reference bounding boxes, evaluated across detection and segmentation 
tasks
\cite{sun2026regionreasoner}. TreeBench and TreeVGR jointly supervise localization 
and reasoning with reinforcement learning, showing that traceable visual evidence 
improves reasoning accuracy \cite{wang2026traceableevidenceenhancedvisual}, while 
MGPO enables models to iteratively focus on key visual regions in high-resolution 
images \cite{huang2025high}. While these methods advance grounded reasoning, they optimize for answer accuracy and do not evaluate whether models identify the complete reasoning evidence required to verify an answer across structured diagram domains. VISTA-Bench reveals that vision-language models perform substantially worse on 
visualized text than on semantically equivalent pure text \cite{liu2026vista}, and  studies on spatial reasoning from language show that models infer certain spatial relationships directly from textual descriptions without visual grounding \cite{guo2026can}. These findings highlight that grounding predictions in the correct visual evidence 
remains an open problem, particularly for structured diagrams where answering questions requires interpreting chart components, labeled regions, connectors, and spatial relationships.

To address these gaps, we introduce \benchmark, a benchmark for 
evidence-grounded reasoning in diagrams. Rather than evaluating answer 
correctness alone, \benchmark evaluates whether models identify the diagram regions 
required to verify an answer. By requiring explicit localization of reasoning 
evidence across multiple diagram domains, \benchmark provides a systematic framework 
for evaluating grounded visual reasoning in diagram question answering.

\section{Our \benchmark{} Benchmark}
\label{sec:dataset}

\begin{wrapfigure}{r}{0.45\textwidth}
    \vspace{-1em}
    \centering
    \includegraphics[width=0.44\textwidth]{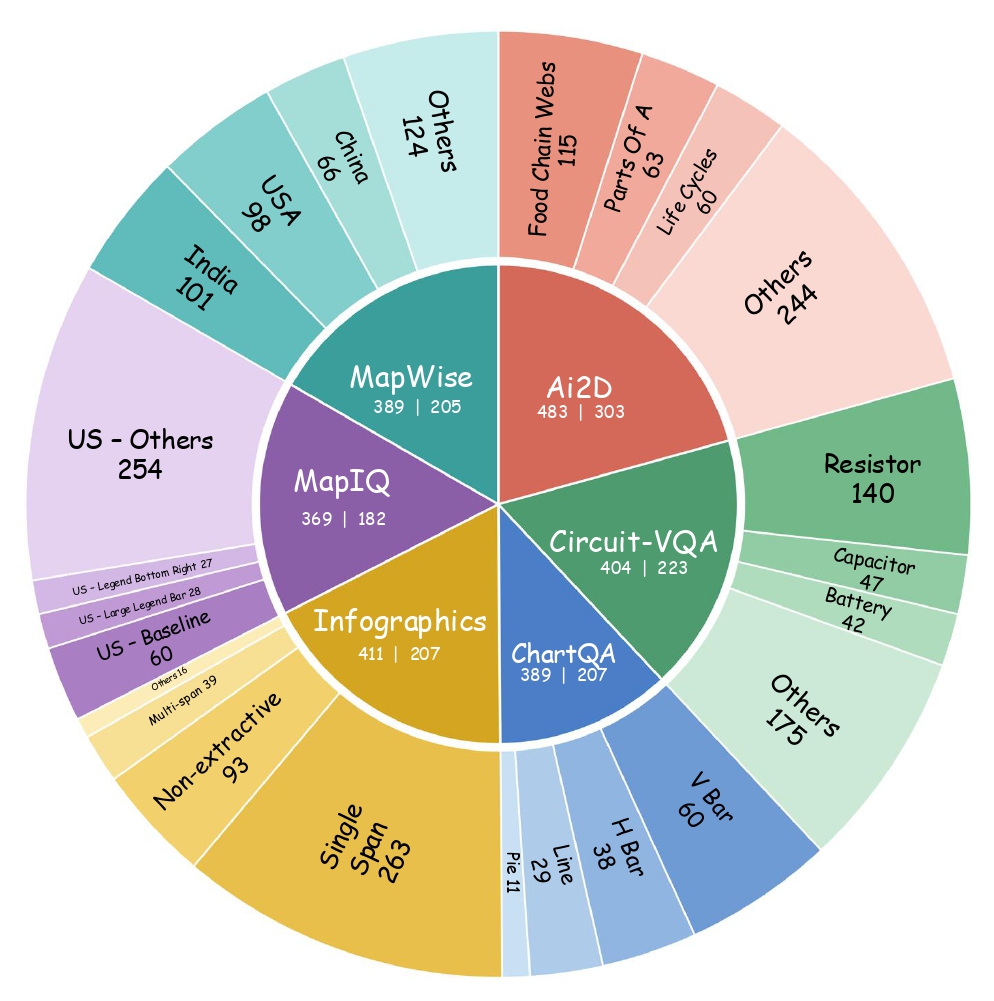}
    \caption{\benchmark benchmark composition.
    \benchmark contains six diagram domains. The inner ring shows instance counts (question count | image count). The outer ring shows sub-category breakdowns reflecting diagram type (e.g., Resistor, V Bar, Food Chain Webs) or question type 
    (e.g., Single Span, Non-extractive) within each domain. \benchmark spans heterogeneous visual structures and reasoning requirements 
    across all six domains.}
    \label{fig:sunburst}
    \vspace{-1.9em}
\end{wrapfigure}

We introduce \benchmark, a human-verified benchmark for evaluating evidence-grounded reasoning in diagrams. We construct the benchmark by combining multiple diagram QA datasets and annotating reasoning-level visual evidence for each instance. We construct the benchmark from six established diagram question answering datasets: AI2D \cite{kembhavi2016diagram}, ChartQA \cite{masry2022chartqa}, Circuit-VQA \cite{mehta2024circuitvqa}, InfographicsVQA \cite{mathew2022infographicvqa}, MapIQ \cite{srivastava2025mapiq}, and MapWise \cite{mukhopadhyay2025mapwise}. These sources cover diverse diagram types, including scientific diagrams, charts, circuit schematics, infographics, and geographic maps. This multi-domain design enables evaluation under heterogeneous visual structures and reasoning requirements.
 
\subsection{Dataset Construction}
\label{subsec:dataset_construction}


We design a unified annotation pipeline to produce high-quality evidence annotations across datasets with heterogeneous annotation formats. The pipeline consists of three stages: (1)~\textbf{Candidate region initialization}, in which we initialize candidate regions from existing 
annotations when available, or via template-based region transfer otherwise; (2)~\textbf{Model-assisted proposal generation}, in which multimodal models 
generate question-conditioned evidence proposals to guide region selection; 
and (3)~\textbf{Human verification and refinement}, in which annotators 
review, refine, and finalize evidence by selecting the minimal set of regions 
required to justify the answer. For each instance, we retain the diagram image $I$, question $q$, and answer $a$, and annotate a gold evidence set $B^\star$ containing the visual regions required for verification. These regions may include answer-bearing elements, labels, legends, or spatially related components, depending on the reasoning requirements. The pipeline operates on dataset records stored in JSON format. When source datasets provide candidate regions, we directly load them into the \diagrams{} interface, where model-generated proposals assist annotators in selecting and refining evidence. When such regions are unavailable, we first construct an initial region pool via template-based transfer, then apply the same proposal generation and human verification workflow.

\begin{figure*}[!t]
    \centering
    \vspace{-1.5em}
    \includegraphics[width=0.9\linewidth]{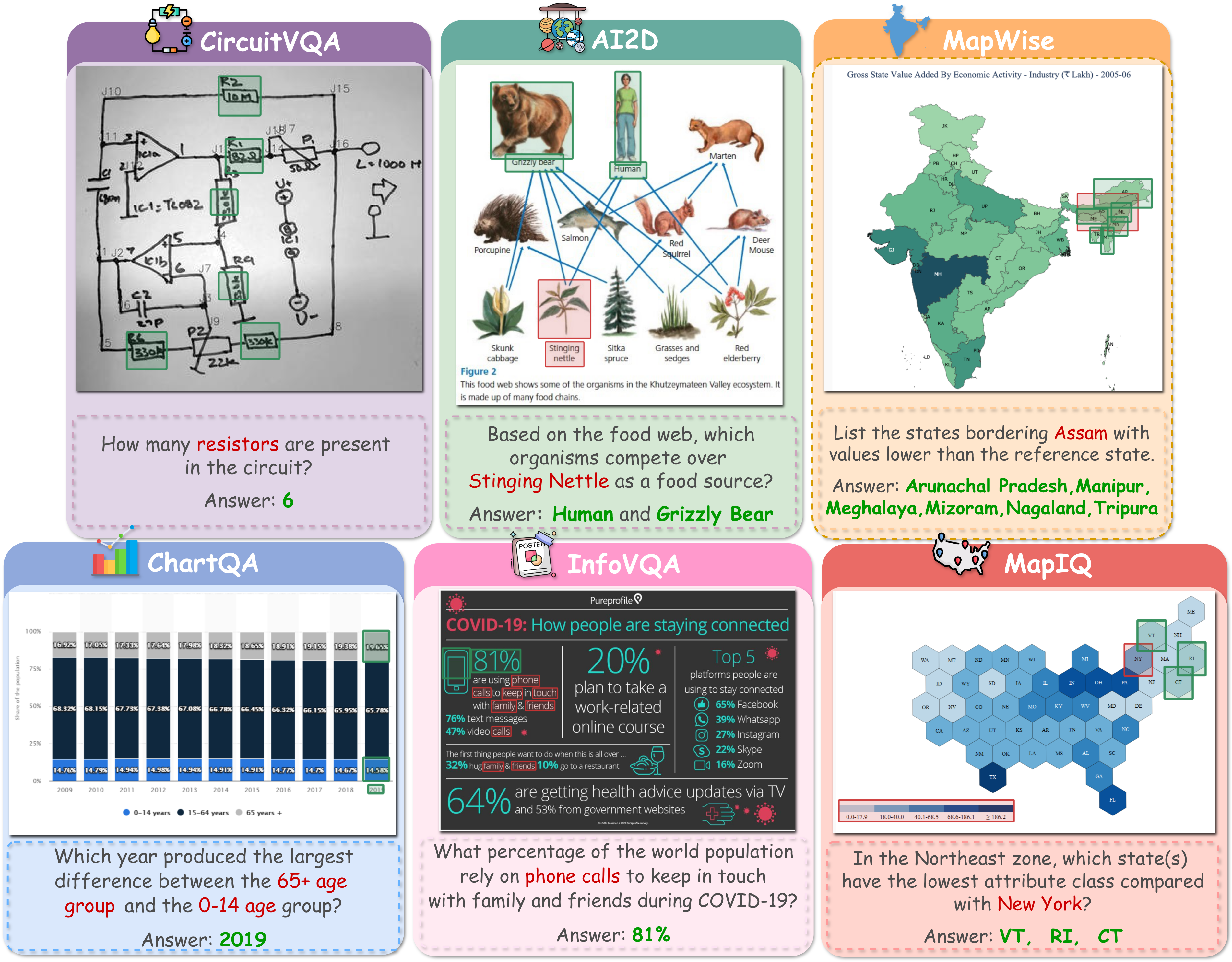}
    \caption{{Examples from the \benchmark Benchmark.} We present illustrative 
examples from each of the six diagram domains, each paired with a question, 
answer, and annotated bounding boxes that identify the visual evidence required 
to justify the answer.}
    \label{fig:main}
\end{figure*}

\begin{figure*}[t]
\centering
\includegraphics[width=1\linewidth]{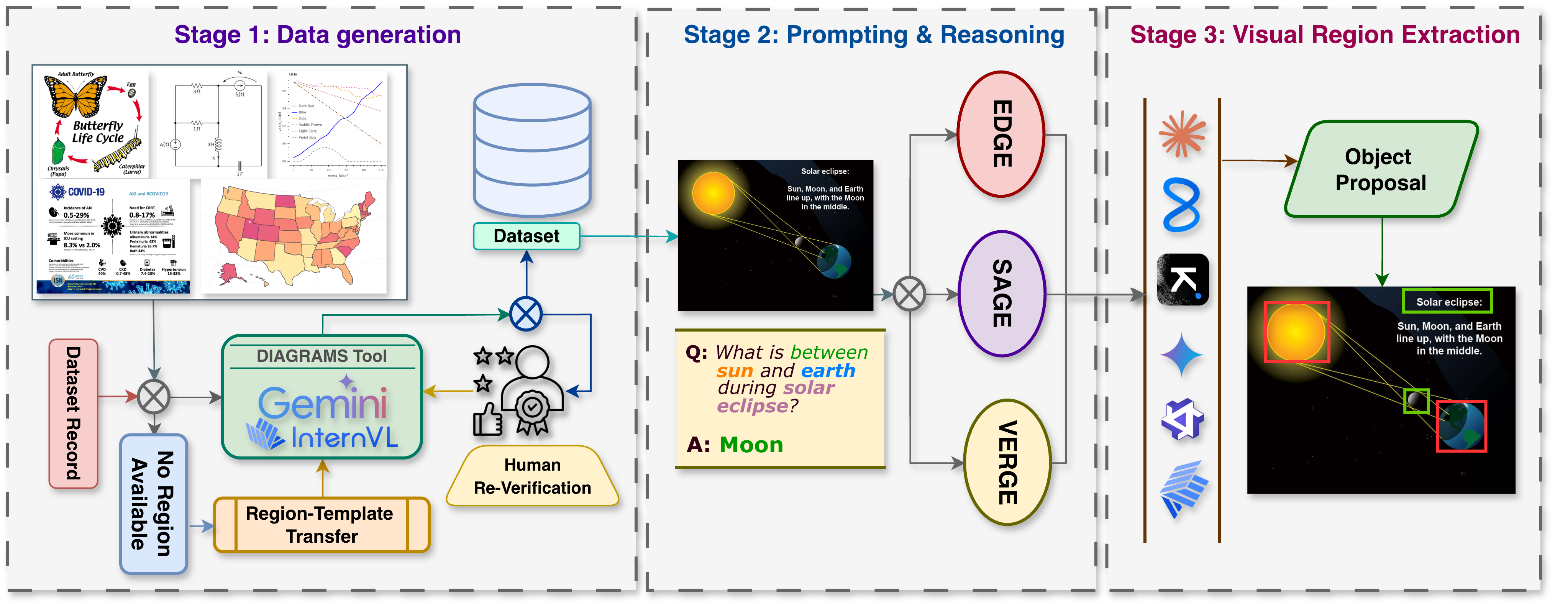}

\caption{Overview of the \benchmark{} benchmark construction and evaluation pipeline. 
\textbf{Stage 1: Data generation.} We collect diagram QA datasets and initialize candidate regions from existing annotations or template-based region transfer. Multimodal models generate candidate evidence, which human annotators verify using the \diagrams{} tool. 
\textbf{Stage 2: Prompting and reasoning.} Vision–language models receive the diagram, question, and answer and produce reasoning-level evidence predictions using EDGE, SAGE, and VERGE prompting strategies. 
\textbf{Stage 3: Visual region extraction.} We convert predicted reasoning elements into bounding boxes and compare them with human-verified annotations to compute grounding metrics.}
\label{fig:overview}
\end{figure*}

This design ensures a consistent annotation process across datasets while reducing annotation effort. The pipeline produces 11,664 annotated question instances across six diagram 
domains. We release the test split of 2,445 human-verified instances together 
with a standardized evaluation framework, and plan to release the training and 
validation splits in future work to support learning methods for grounded 
diagram reasoning. Figure~\ref{fig:sunburst} summarizes the benchmark composition, Figure~\ref{fig:main} presents illustrative examples from each domain, 
eand Figure~\ref{fig:overview} illustrates the overall pipeline.

\subsubsection{Template-Based Region Transfer}
\label{subsec:template_transfer}

For datasets without region annotations, we initialize candidate regions using geometric alignment between template and target diagrams.

A human first creates a template diagram and annotates its regions with bounding boxes. Let the template image be $I_t$ and the target image be $I_s$. Let $B_t = \{b_1, \ldots, b_n\}$ denote the set of template bounding boxes, where each box $b_i$ is represented by its four corner points. Our objective is to estimate corresponding regions $B_s$ on the target image. This procedure provides a structured initialization that accelerates annotation while preserving spatial consistency.


We estimate a geometric transformation between the template and target images through feature matching followed by homography estimation \cite{lowe2004distinctive,hartley2003multiple}. Given matched keypoints $\{x_i\}$ in $I_t$ and $\{x_i'\}$ in $I_s$, we compute a homography matrix $H$ such that $ x_i' \sim Hx_i.$ We estimate $H$ with RANSAC to reject mismatched correspondences \cite{fischler1981random}. For each template box $b_i$ with corner set $C_i = \{p_1, p_2, p_3, p_4\}$, we transform its corners into the target image: \[
C_i' = \{Hp_1, Hp_2, Hp_3, Hp_4\}.
\] We then convert the transformed quadrilateral into an axis-aligned bounding box: 
$
b_i' =
\left[
\min_x(C_i'),\;
\min_y(C_i'),\;
\max_x(C_i'),\;
\max_y(C_i')
\right].
$ When homography estimation fails due to insufficient feature correspondences, we instead apply proportional scaling. Let the template image dimensions be $(W_t, H_t)$ and the target image dimensions be $(W_s, H_s)$. We compute the scale factors \[
s_x = \frac{W_s}{W_t}, \quad s_y = \frac{H_s}{H_t},
\]
and scale the template box coordinates accordingly.
Annotators subsequently verify, edit, discard, or supplement these regions during the review stage. The implementation follows standard computer vision routines provided by the OpenCV library \cite{bradski2000opencv}.

\subsubsection{Human Verification}
\label{subsec:human_verification}

After generating candidate regions, annotators verify evidence selections in the \diagrams{} interface. For each instance, annotators inspect the diagram together with the question and answer and select the minimal set of visual regions required to justify the answer. During review, annotators may retain correct proposals, refine imprecise regions, remove irrelevant selections, or add missing evidence when necessary. This process produces the final gold evidence set $B^\star$ for each instance. Six annotators participated in the verification process under a shared protocol. Appendix~\ref{appendix:annotation_instructions} summarizes the annotation instructions, and Appendix~\ref{appendix:iaa} presents the agreement criteria in detail. 

\begin{wraptable}{r}{0.55\textwidth}
    \vspace{-2.0em}
    \centering
    \small
    \caption{Inter-annotator agreement for reasoning-level evidence selection under Complete Visual Reasoning (CVR) and Core Evidence Alignment (CEA). See Appendix~\ref{appendix:iaa} for definitions and evaluation details.}
    \begin{tabular}{llcc}
    \toprule
    \textbf{Dataset} & \textbf{Criterion} & \textbf{Agreement (\%)} & \textbf{$\kappa$} \\
    \midrule
ChartQA     & CVR & 87.0 & 0.658 \\
ChartQA     & CEA & 83.0 & 0.589 \\
Circuit-VQA & CVR & 73.0 & 0.464 \\
Circuit-VQA & CEA & 85.0 & 0.540 \\
\bottomrule
\end{tabular}
\label{tab:iaa_main}
    \vspace{-0.5em}
\end{wraptable}

\subsubsection{Inter-Annotator Agreement}
\label{subsubsec:iaa_main}

We assess annotation reliability on a subset of 100 instances per dataset for the selection-based proposal workflow. Table~\ref{tab:iaa_main} reports agreement scores under both criteria for ChartQA and Circuit-VQA. We evaluate agreement under two complementary criteria: \emph{Complete Visual Reasoning} (CVR), which requires full coverage of the visual reasoning chain, and \emph{Core Evidence Alignment} (CEA), which requires correct grounding of the primary evidence elements. Appendix~\ref{appendix:iaa} defines both criteria and describes the evaluation scope.

\section{Modeling Approaches}
\label{sec:modeling}
We evaluate evidence-grounded reasoning using three prompting-based evaluation protocols: \textbf{EDGE}, \textbf{SAGE}, and \textbf{VERGE}. Each protocol takes as input a diagram image $I$, question $q$, and verified answer $a$, and requires the model to predict a set of bounding boxes corresponding to the visual evidence needed to justify the answer. Formally, each protocol produces a predicted evidence set
\[
\hat{B} = \{(x_k, y_k, w_k, h_k)\}_{k=1}^{K},
\]
where each tuple denotes a bounding box in pixel space. The objective is to approximate the gold evidence set $B^\star$ defined in Section~3. The protocols differ in how they structure the reasoning and grounding process, enabling us to analyze how decomposition and verification affect evidence localization.

\subsection{EDGE: Direct Evidence Grounding}

EDGE performs single-step grounding without explicit intermediate reasoning. Given $(I, q, a)$, the model directly predicts the full evidence set: $\hat{B} = f_{\text{EDGE}}(I, q, a)$. The prompt provides the diagram, question, answer, and an output format specification. The model returns bounding boxes corresponding to all regions required for reasoning, including answer-bearing elements and supporting structures. EDGE serves as a minimal baseline that evaluates whether models can directly ground reasoning-level evidence without decomposition.

\subsection{SAGE: Select-and-Ground Evidence}

SAGE decomposes grounding into two sequential stages: evidence selection and spatial grounding. In the first stage, the model predicts a structured set of evidence elements $E = \{d_j\}_{j=1}^{M}$,
where each $d_j$ is a textual description of a relevant visual element. We denote this stage as
$E = f_{\text{select}}(I, q, a)$. In the second 
stage, the model grounds each element spatially:
\begin{equation}
    \hat{B} = f_{\text{ground}}(I, q, a, E).
    \label{eq:sage}
\end{equation}
This decomposition separates \emph{what to look for} from \emph{where to 
look}, allowing us to evaluate whether explicit evidence planning improves 
grounding quality.

\subsection{VERGE: Verify-and-Refine Grounding}

VERGE extends EDGE with an explicit self-verification stage. The model first 
produces an initial estimate $\hat{B}^{(0)} = f_{\text{EDGE}}(I, q, a)$, 
then refines it by re-examining the diagram:
\begin{equation}
    \hat{B}^{(1)} = f_{\text{verify}}(I, q, a, \hat{B}^{(0)}).
    \label{eq:verge}
\end{equation}
During verification the model adjusts inaccurate boxes, adds missing evidence 
regions, and refines spatial coverage, while the prompt constrains updates to 
preserve valid predictions. 

Together, EDGE, SAGE, and VERGE define a spectrum of grounding protocols ranging from direct prediction to structured reasoning and iterative refinement. This design enables controlled analysis of how reasoning structure influences evidence localization, without requiring model retraining. Appendix~\ref{app:prompts} provides the complete prompt templates for 
all three strategies.

\section{Experimental Setup}
\label{sec:exp}

We evaluate \benchmark using eight recent vision-language models spanning both proprietary and open-weight systems. Our model suite includes Claude Opus 4.6 \cite{Anthro2024Claude}, Claude Sonnet 4.6, Kimi K2.5 \cite{Moonshot2026Kimi}, Gemini 3 Pro \cite{Google2026Gemini3}, and Llama 4 Maverick \cite{Meta2025Llama4} as API-based models, and Qwen3.5-35B-A3B \cite{Alibaba2025Qwen3}, Gemma3-27B-IT \cite{Google2025Gemma3}, and InternVL3.5-38B \cite{Wang2025InternVL35} as locally hosted open-weight models. This set covers a broad range of multimodal architectures, deployment settings, and visual reasoning capabilities. We use the publicly available inference interfaces or released checkpoints for all models and do not apply additional fine-tuning. Appendix~\ref{app:exp_details} provides the full model identifiers, decoding configuration, image pre-processing settings, and infrastructure details.

We evaluate model outputs using complementary metrics that capture both localization quality and evidence coverage. Specifically, we report Max Pairwise IoU \cite{lin2014microsoft}, Grounding IoU, threshold hit rates derived from these measures, and box-level precision, recall, and F1. 

Appendix~\ref{app:exp_details} provides full metric definitions. 
Tables~\ref{tab:threshold_table_0.3} and~\ref{tab:threshold_table_0.5} 
report thresholded results at $\tau{=}0.3$ and $\tau{=}0.5$ respectively, 
showing how localization quality degrades under stricter alignment 
requirements.

\begin{table*}[t]
\caption{Prompt-level grounding performance across models and diagram domains. We compare three prompting strategies (EDGE, SAGE, VERGE) using Max-Pair IoU (MP$_{\mathrm{IoU}}$), Grounding IoU (G$_{\mathrm{IoU}}$), and F1.The table shows strong prompt sensitivity and a consistent gap between coarse localization and reasoning-level evidence grounding across datasets.}
\centering
\small
\setlength{\tabcolsep}{3pt}
\renewcommand{\arraystretch}{1.1}

\resizebox{\textwidth}{!}{

\begin{tabular}{l|ccc|ccc|ccc|ccc|ccc|ccc}

\toprule
& \multicolumn{3}{c|}{\textbf{ChartQA}} 
& \multicolumn{3}{c|}{\textbf{Circuit-VQA}} 
& \multicolumn{3}{c|}{\textbf{InfoVQA}} 
& \multicolumn{3}{c|}{\textbf{MapIQ}} 
& \multicolumn{3}{c|}{\textbf{MapWise}} 
& \multicolumn{3}{c}{\textbf{AI2D}} \\

\cmidrule(lr){2-4}
\cmidrule(lr){5-7}
\cmidrule(lr){8-10}
\cmidrule(lr){11-13}
\cmidrule(lr){14-16}
\cmidrule(lr){17-19}

\textbf{Prompt}
& \makecell{\textbf{MP$_{\mathrm{IoU}}$}} & \makecell{\textbf{G$_{\mathrm{IoU}}$}} & \textbf{F1}
& \makecell{\textbf{MP$_{\mathrm{IoU}}$}} & \makecell{\textbf{G$_{\mathrm{IoU}}$}} & \textbf{F1}
& \makecell{\textbf{MP$_{\mathrm{IoU}}$}} & \makecell{\textbf{G$_{\mathrm{IoU}}$}} & \textbf{F1}
& \makecell{\textbf{MP$_{\mathrm{IoU}}$}} & \makecell{\textbf{G$_{\mathrm{IoU}}$}} & \textbf{F1}
& \makecell{\textbf{MP$_{\mathrm{IoU}}$}} & \makecell{\textbf{G$_{\mathrm{IoU}}$}} & \textbf{F1}
& \makecell{\textbf{MP$_{\mathrm{IoU}}$}} & \makecell{\textbf{G$_{\mathrm{IoU}}$}} & \textbf{F1} \\

\hline

\multicolumn{19}{>{\columncolor{lightestgray}}c}{\textbf{Claude Opus 4.6}} \\
\hline
\textbf{EDGE}  & 27.6 & 13.9 & 9.0 & 7.0 & 4.1 & 3.5 & 8.7 & 4.5 & 3.6 & 51.1 & \textbf{31.0} & \textbf{22.9} & 20.4 & 8.5 & 7.8 & 34.5 & 23.7 & \textbf{15.8} \\
\textbf{SAGE}  & 8.6  & 4.6  & 2.4 & 12.2 & 3.8 & 3.0 & \textbf{12.5} & 4.1 & 3.7 & 51.6 & 24.3 & 20.3 & 19.4 & 7.0 & 4.8 & \textbf{42.5} & \textbf{24.2} & 14.1 \\
\textbf{VERGE} & \textbf{35.7} & 15.6 & 9.3 & 10.3 & 3.8 & 3.6 & 11.0 & 4.2 & 3.7 & \textbf{58.4} & 30.0 & 22.3 & 23.9 & 9.5 & 6.5 & 40.6 & 23.5 & 14.3 \\

\hline
\multicolumn{19}{>{\columncolor{lightestgray}}c}{\textbf{Kimi K2.5}} \\
\hline
\textbf{EDGE}  & 32.1 & \textbf{18.6}& \textbf{11.8} & 7.5 & 4.1 & 3.6 & 10.3 & 4.9 & \textbf{4.9} & 36.5 & 17.4 & 13.6 & 23.8 & 9.4 & \textbf{10.1} & 30.4 & 18.3 & 13.8 \\
\textbf{SAGE}  & 32.6 & 13.7 & 7.9 & \textbf{14.3} & 2.8 & 2.0 & 10.1 & 2.1 & 2.1 & 46.4 & 18.3 & 12.8 & \textbf{37.3} & \textbf{13.3} & 9.7 & 42.2 & 20.9 & 12.4 \\
\textbf{VERGE} & 34.6 & 13.0 & 8.7 & 11.6 & 3.4 & 2.7 & 9.6  & 3.0 & 3.1 & 40.2 & 16.6 & 11.2 & 26.6 & 9.7  & 8.0 & 35.1 & 17.7 & 10.3 \\

\hline
\multicolumn{19}{>{\columncolor{lightestgray}}c}{\textbf{Gemini 3 Pro}} \\
\hline
\textbf{EDGE}  & 16.1 & 9.7 & 6.5 & 10.0 & \textbf{5.8} & \textbf{5.8} & 8.8  & 5.3 & 4.3 & 14.6 & 6.1 & 4.0 & 19.4 & 11.3 & 9.4 & 29.3 & 18.4 & 14.9 \\
\textbf{SAGE} & 22.3 & 14.0 & 9.2 & 9.1  & 5.4 & 5.6 & 10.9 & \textbf{7.3} & 4.8 & 10.1 & 2.8 & 3.1 & 9.7  & 5.0 & 5.0 & 20.2 & 11.0 & 11.0 \\
\textbf{VERGE} & 9.4  & 5.4 & 3.7 & 9.2  & 5.7 & \textbf{5.8} & 6.3  & 3.8 & 3.0 & 8.0  & 4.1 & 2.3 & 15.6 & 9.3 & 8.6 & 20.9 & 13.3 & 11.9 \\

\hline
\multicolumn{19}{>{\columncolor{lightestgray}}c}{\textbf{InternVL3.5-38B}} \\
\hline
\textbf{EDGE}  & 5.5 & 3.5 & 2.0 & 2.5 & 1.8 & 1.5 & 1.3 & 0.6 & 0.5 & 8.7 & 1.8 & 2.0 & 5.1 & 2.7 & 2.3 & 9.3 & 6.6 & 4.9 \\
\textbf{SAGE}  & 10.5 & 3.8 & 3.0 & 7.3 & 2.2 & 1.9 & 3.9 & 1.7 & 1.1 & 7.1 & 3.6 & 2.0 & 7.6 & 2.8 & 2.5 & 19.3 & 11.1 & 6.8 \\
\textbf{VERGE} & 8.6 & 4.1 & 2.8 & 5.6 & 2.1 & 2.0 & 2.1 & 1.0 & 0.7 & 10.1 & 2.1 & 2.1 & 5.2 & 2.5 & 2.1 & 16.4 & 10.0 & 6.3 \\
\bottomrule

\end{tabular}
}

\label{tab:prompt_dataset_model}

\end{table*}

\section{Results and Analysis}
 \label{sec:results}
We analyze grounding performance through five questions: how well current models ground reasoning evidence, how closed-source and open-weight models differ, how prompting strategies affect grounding, whether models localize relevant regions without capturing complete evidence, and which diagram domains remain most difficult. We answer these questions using the prompt-level results in Table~\ref{tab:prompt_dataset_model} and the model-aggregated results in Table~\ref{tab:dataset_model}.

\subsection{Evidence Grounding Capability}

\paragraph{Do current vision–language models achieve strong evidence grounding across diagram domains?}  No. Table~\ref{tab:dataset_model} shows that grounding remains weak across most domains: the best F1 reaches only 9.5 on ChartQA, 5.7 on Circuit-VQA, 4.0 on InfoVQA, and 9.3 on MapWise. Only MapIQ and AI2D show clearly stronger results, where Claude Opus~4.6 reaches 21.8 and 14.7 F1, respectively. These results show that current models can often find relevant evidence but rarely ground complete reasoning chains across domains.
\vspace{-6pt}
\noindent \paragraph{Which models show the strongest grounding capability?}
Performance varies by domain rather than following a single overall ranking. Table~\ref{tab:dataset_model} shows that Claude Opus~4.6 achieves the strongest results on MapIQ with 53.7 MP$_{\mathrm{IoU}}$, 28.4 G$_{\mathrm{IoU}}$, and 21.8 F1, and on AI2D with 39.2 MP$_{\mathrm{IoU}}$, 23.8 G$_{\mathrm{IoU}}$, and 14.7 F1. Kimi~K2.5 achieves the best aggregated performance on ChartQA and MapWise, reaching 9.5 and 9.3 F1, while Gemini~3~Pro achieves the best aggregated results on Circuit-VQA and InfoVQA with 5.7 and 4.0 F1. Table~\ref{tab:prompt_dataset_model} shows the same domain-specific pattern across prompting strategies.
\vspace{-6pt}
\noindent\paragraph{Does grounding capability remain stable across domains and prompts?}
No. Table~\ref{tab:prompt_dataset_model} shows large within-model variation across prompts: Claude Opus~4.6 on ChartQA rises from 2.4 F1 under SAGE to 9.3 under VERGE, while Gemini~3~Pro on AI2D ranges from 11.0 under SAGE to 14.9 under EDGE. Table~\ref{tab:dataset_model} shows equally large cross-domain variation: Claude Opus~4.6 reaches 21.8 F1 on MapIQ but only 3.4 on Circuit-VQA. Grounding performance, therefore, depends strongly on both prompt design and diagram domain. We present qualitative success and failure cases across all eight models 
and all three prompting strategies in Appendix~\ref{appendix:qualitative}, 
including model-wise comparisons on MapWise (Figure~\ref{fig:model_success}), 
ChartQA (Figure~\ref{fig:model_failure}), and prompt-wise comparisons 
across all six domains (Figures~\ref{fig:prompt_success} 
and~\ref{fig:prompt_failure}).

\subsection{Open-Weight vs Closed-Source Models}

\paragraph{Do closed-source models outperform open-weight models?}
Yes. Table~\ref{tab:dataset_model} shows a clear gap between the two groups across all datasets. On MapIQ, Claude Opus~4.6 reaches 21.8 F1, while the best open-weight model, InternVL3.5-38B, reaches only 2.0. On ChartQA, the strongest open-weight model (Kimi~K2.5) reaches 9.5 F1, while other open-weight models remain substantially lower (e.g., 2.6 from InternVL3.5-38B). This gap persists across MapWise and AI2D.

\vspace{-6pt}
\noindent\paragraph{Does prompting close the gap between closed-source and open-weight models?}
No. Table~\ref{tab:prompt_dataset_model} shows that prompting improves both groups, but preserves the ranking. Claude Opus~4.6 reaches 22.9 F1 on MapIQ under EDGE and 15.8 on AI2D, while InternVL3.5-38B peaks at only 2.1 on MapIQ and 6.8 on AI2D. Prompting, therefore, improves performance but does not eliminate the gap between model families.

\begin{table*}[t]
\caption{Model-level grounding performance across diagram domains averaged over EDGE, SAGE, and VERGE prompting. We report Max-Pair IoU (MP$_{\mathrm{IoU}}$), Grounding IoU (G$_{\mathrm{IoU}}$), and F1. Across models, MP$_{\mathrm{IoU}}$ consistently exceeds G$_{\mathrm{IoU}}$ and F1, indicating that models often localize approximate evidence regions but fail to capture the complete reasoning evidence required for diagram question answering.}
\centering
\small
\setlength{\tabcolsep}{3pt}
\renewcommand{\arraystretch}{1.3}

\resizebox{\textwidth}{!}{
\begin{tabular}{l|ccc|ccc|ccc|ccc|ccc|ccc}

\toprule
& \multicolumn{3}{c|}{\textbf{ChartQA}} 
& \multicolumn{3}{c|}{\textbf{Circuit-VQA}} 
& \multicolumn{3}{c|}{\textbf{InfoVQA}} 
& \multicolumn{3}{c|}{\textbf{MapIQ}} 
& \multicolumn{3}{c|}{\textbf{MapWise}} 
& \multicolumn{3}{c}{\textbf{AI2D}} \\

\cmidrule(lr){2-4}
\cmidrule(lr){5-7}
\cmidrule(lr){8-10}
\cmidrule(lr){11-13}
\cmidrule(lr){14-16}
\cmidrule(lr){17-19}

\textbf{Model}
& \makecell{\textbf{MP$_{\mathrm{IoU}}$}} & \makecell{\textbf{G$_{\mathrm{IoU}}$}} & \textbf{F1}
& \makecell{\textbf{MP$_{\mathrm{IoU}}$}} & \makecell{\textbf{G$_{\mathrm{IoU}}$}} & \textbf{F1}
& \makecell{\textbf{MP$_{\mathrm{IoU}}$}} & \makecell{\textbf{G$_{\mathrm{IoU}}$}} & \textbf{F1}
& \makecell{\textbf{MP$_{\mathrm{IoU}}$}} & \makecell{\textbf{G$_{\mathrm{IoU}}$}} & \textbf{F1}
& \makecell{\textbf{MP$_{\mathrm{IoU}}$}} & \makecell{\textbf{G$_{\mathrm{IoU}}$}} & \textbf{F1}
& \makecell{\textbf{MP$_{\mathrm{IoU}}$}} & \makecell{\textbf{G$_{\mathrm{IoU}}$}} & \textbf{F1} \\

\midrule

\textbf{Claude Opus 4.6}
& 23.9 & 11.4 & 6.9
& 9.8 & 3.9 & 3.4
& \textbf{10.7} & 4.3 & 3.7
& \textbf{53.7} & \textbf{28.4} & \textbf{21.8}
& 21.2 & 8.3 & 6.4
& \textbf{39.2} & \textbf{23.8} & \textbf{14.7} \\

\textbf{Claude Sonnet 4.6}
& 13.7 & 7.7 & 2.3
& 4.4 & 2.2 & 0.1
& 3.2 & 1.6 & 0.1
& 40.6 & 21.3 & 13.8
& 23.9 & 9.7 & 4.2
& 23.8 & 15.9 & 5.5 \\

\textbf{Kimi K2.5}
& \textbf{33.1} & \textbf{15.1} & \textbf{9.5}
& \textbf{11.1} & 3.4 & 2.8
& 10.0 & 3.3 & 3.4
& 41.0 & 17.4 & 12.5
& \textbf{29.2} & \textbf{10.8} & \textbf{9.3}
& 35.9 & 19.0 & 12.2 \\

\textbf{Gemini 3 Pro}
& 15.9 & 9.7 & 6.5
& 9.4 & \textbf{5.6} & \textbf{5.7}
& 8.7 & \textbf{5.5} & \textbf{4.0}
& 10.9 & 4.3 & 3.1
& 14.9 & 8.5 & 7.7
& 23.5 & 14.2 & 12.6 \\
\textbf{Llama4-Maverick-17B-IT}
& 7.7 & 4.9 & 0.6
& 3.6 & 1.8 & 0.2
& 1.9 & 1.5 & 0.0
& 15.7 & 6.2 & 1.0
& 11.1 & 5.2 & 1.1
& 13.2 & 9.5 & 2.4 \\

\textbf{Qwen3.5-3.5B-A3B}
& 12.6 & 5.7 & 0.8
& 5.4 & 2.1 & 0.2
& 2.6 & 1.3 & 0.0
& 18.3 & 4.7 & 0.8
& 12.3 & 4.3 & 0.8
& 16.3 & 9.2 & 1.2 \\

\textbf{Gemma3-27B-IT}
& 8.0 & 4.1 & 0.2
& 6.6 & 1.6 & 0.2
& 4.2 & 1.1 & 0.1
& 18.8 & 4.5 & 0.7
& 15.2 & 5.1 & 0.7
& 19.0 & 12.4 & 2.0 \\

\textbf{InternVL3.5-38B}
& 8.2 & 3.8 & 2.6
& 5.1 & 2.0 & 1.8
& 2.4 & 1.1 & 0.8
& 8.6 & 2.5 & 2.0
& 6.0 & 2.7 & 2.3
& 15.0 & 9.2 & 6.0 \\

\bottomrule
\end{tabular}
}

\label{tab:dataset_model}
\end{table*}

\subsection{Effect of Prompting Strategies}

\paragraph{Do prompting strategies materially affect grounding performance?}
Yes. Table~\ref{tab:prompt_dataset_model} shows substantial prompt sensitivity across models and datasets. On ChartQA, Claude Opus~4.6 rises from 2.4 F1 under SAGE to 9.3 under VERGE, while Kimi~K2.5 reaches its best result under EDGE with 11.8 F1. Similar shifts appear on AI2D, where Gemini~3~Pro drops from 14.9 under EDGE to 11.0 under SAGE. Prompt design clearly changes grounding outcomes. Table~\ref{tab:prompt_dataset_model_others} in the appendix extends this 
analysis to the remaining four models, showing that prompting effects remain 
highly inconsistent and that weaker models produce non-monotonic behavior 
across EDGE, SAGE, and VERGE.
\vspace{-6pt}
\noindent\paragraph{Does one prompt dominate across all settings?}
No. EDGE, SAGE, and VERGE each help different settings. Table~\ref{tab:prompt_dataset_model} shows that VERGE gives Claude Opus~4.6 its best ChartQA and MapIQ localization scores at 35.7 and 58.4 MP$_{\mathrm{IoU}}$, EDGE gives Kimi~K2.5 its best ChartQA and MapWise F1 at 11.8 and 10.1, and SAGE gives Kimi~K2.5 its best Circuit-VQA localization at 14.3 MP$_{\mathrm{IoU}}$. No single prompting strategy dominates across tasks.
\vspace{-6pt}
\noindent\paragraph{What aspect of grounding benefits most from prompting?}
Prompting helps localization more than complete evidence coverage. Table~\ref{tab:prompt_dataset_model} shows that MP$_{\mathrm{IoU}}$ often shifts more than F1 across prompts. For example, Claude Opus~4.6 on MapIQ increases from 51.1 to 58.4 MP$_{\mathrm{IoU}}$ between EDGE and VERGE, but F1 changes only from 22.9 to 22.3. Prompting therefore helps models find relevant regions more reliably than it helps them recover full reasoning evidence.

\subsection{Localization vs Complete Evidence}

\paragraph{Do models localize relevant regions better than they ground complete evidence?}
Yes. Table~\ref{tab:dataset_model} shows a consistent gap between localization and full grounding. Claude Opus~4.6 reaches 53.7 MP$_{\mathrm{IoU}}$ on MapIQ but only 28.4 G$_{\mathrm{IoU}}$ and 21.8 F1, while Kimi~K2.5 reaches 33.1, 15.1, and 9.5 on ChartQA. These gaps show that models often identify where evidence lies without fully capturing all regions needed for complete reasoning support.

\vspace{-6pt}
\noindent\paragraph{Does this localization-coverage gap persist across models?}
Yes. The gap appears across both strong and weak systems. Table~\ref{tab:dataset_model} shows that Gemini~3~Pro on AI2D achieves 23.5 MP$_{\mathrm{IoU}}$, 14.2 G$_{\mathrm{IoU}}$, and 12.6 F1, while InternVL3.5-38B achieves 15.0, 9.2, and 6.0. The same ordering appears in Table~\ref{tab:prompt_dataset_model}, which shows that prompt changes do not remove the gap. This pattern indicates a systematic limitation in evidence coverage.
\vspace{-6pt}
\noindent\paragraph{Do prompts reduce the localization-coverage gap?} Only slightly. Table~\ref{tab:prompt_dataset_model} shows that prompts can improve MP$_{\mathrm{IoU}}$ without producing comparable gains in F1. For Claude Opus~4.6 on MapIQ, VERGE raises MP$_{\mathrm{IoU}}$ to 58.4 from 51.1 under EDGE, but F1 stays nearly flat at 22.3 versus 22.9. Prompting therefore improves region finding more than evidence completeness.

\subsection{Domain Difficulty}
\paragraph{Which domains are hardest for evidence grounding?}
Circuit-VQA and InfoVQA are the hardest domains. Table~\ref{tab:dataset_model} shows that the best Circuit-VQA F1 reaches only 5.7 from Gemini~3~Pro, and the best InfoVQA F1 reaches only 4.0, again from Gemini~3~Pro. Most other models stay below 3.5 on both datasets. These results suggest that dense structure, relational dependencies, and visually crowded layouts sharply increase grounding difficulty.

\vspace{-6pt}
\noindent\paragraph{How does grounding performance vary across diagram domains?}
Table~\ref{tab:dataset_model} shows that models obtain relatively higher scores on MapIQ and AI2D than on other domains. For example, Claude Opus~4.6 reaches 21.8 F1 on MapIQ and 14.7 on AI2D, while Kimi~K2.5 reaches 12.5 and 12.2. Table~\ref{tab:prompt_dataset_model} shows a similar trend across prompting strategies, with several models exceeding 20 G$_{\mathrm{IoU}}$ or achieving double-digit F1 on these datasets. Clearer spatial layouts and explicit labels likely make evidence localization comparatively easier, although  performance remains low overall.

\section{Conclusion}
We introduce \benchmark, a benchmark for evidence-grounded visual reasoning in diagram question answering. Existing diagram QA benchmarks measure only answer correctness, which cannot distinguish models that genuinely ground their reasoning in the diagram from those that exploit textual priors or 
dataset artifacts.\benchmark addresses this limitation by requiring models 
to predict bounding boxes that identify the visual regions justifying a given 
answer, enabling direct evaluation of reasoning faithfulness rather than answer 
prediction alone. We construct the benchmark from six diagram QA datasets 
spanning charts, scientific diagrams, maps, circuit schematics, and 
infographics, and we produce 2,445 human-verified test instances through a 
three-stage pipeline combining model-assisted proposals with multi-annotator 
verification. The resulting benchmark provides a curated test set together with a standardized evaluation protocol that measures reasoning-level grounding quality. We evaluate eight recent vision-language models under three prompting 
strategies (EDGE, SAGE, and VERGE) and find a consistent gap between answer accuracy and evidence grounding performance: models frequently produce correct answers 
while failing to identify the diagram regions that support those answers, 
revealing a fundamental limitation of current multimodal reasoning systems 
that answer-only evaluation cannot expose. This gap persists across 
prompt designs, model families, and diagram domains, revealing a limitation 
of current multimodal reasoning systems that answer-only evaluation cannot 
expose. We 
release the dataset, annotations, and evaluation code to support future work 
on faithful diagram reasoning, grounding-aware architectures, and 
training methods such as reinforcement learning methods that reward 
models for producing visual evidence aligned with human verification that 
targets reasoning faithfulness rather than answer accuracy alone. We discuss limitations of the bounding-box representation and annotation 
subjectivity in Appendix~\ref{app:limitations}.


\bibliographystyle{unsrtnat}
\bibliography{custom}





\clearpage
\appendix



\section{Limitations}
\label{app:limitations}
\benchmark represents visual evidence using rectangular bounding boxes, which enables efficient annotation and consistent evaluation across all six diagram domains but provides only coarse spatial localization. Bounding boxes may enclose surrounding pixels that do not belong to the minimal evidence set, particularly for thin or irregular structures such as wires, curved 
connectors, and long directional arrows. In addition, identifying the minimal sufficient visual evidence required to justify an answer can involve subjective judgment especially for questions requiring multi-hop reasoning across spatially separated regions, where different annotators may select evidence sets that are distinct but equally valid. We mitigate this variability through detailed annotation guidelines and a multi-stage human verification process, but minor differences in evidence selection may still occur. Finally, the current release focuses on a standardized evaluation test set and accompanying evaluation framework to encourage consistent benchmarking across models. While this design promotes fair comparison, it limits immediate supervised training directly on \benchmark. The current release focuses on the test split to encourage consistent 
benchmarking across models. We plan to release the training and validation 
splits in future work to support supervised training for grounded diagram 
reasoning.

\section{Ethics Statement}
\label{app:ethics}
We construct DRAGON using publicly available
diagram question answering datasets, including AI2D, ChartQA, Circuit-VQA, InfographicsVQA, MapIQ, and 
MapWise, and we follow their respective licenses. We distribute ChartQA, 
Circuit-VQA, MapIQ, and MapWise directly under their CC BY 4.0 licenses. 
We do not redistribute AI2D images, which carry a non-commercial license from 
the Allen Institute for Artificial Intelligence, or InfographicsVQA images, 
which require separate download from the DocVQA challenge under their 
respective terms. Users must download both datasets independently following 
the instructions in our dataset README. Our release includes question-answer 
pairs and our derived evidence annotations integrated into a unified JSON 
schema. We partition the full 11,664-instance dataset into training, 
validation, and test splits, releasing only the 2,445-instance test split 
at this time.

We design the annotation workflow to promote consistency and transparency 
through standardized guidelines and multi-stage human verification. 
Nevertheless, identifying minimal reasoning evidence may still involve 
subjective decisions, particularly for multi-step reasoning tasks. Because 
\benchmark builds on existing datasets, it may inherit biases such as domain 
imbalance or geographic skew present in the original sources. Evaluating large 
vision-language models also incurs computational cost. We mitigate this by 
providing a fixed evaluation split and encourage responsible use of the 
benchmark as a research resource rather than a production tool. We also 
employed AI tools, including large language models, to assist with aspects of 
the project such as prompt development and explanatory generation. All 
AI-generated outputs were reviewed and refined by human authors to ensure 
accuracy and clarity. Overall, this project reflects our commitment to data 
privacy, transparency, annotator welfare, and the responsible integration of 
AI tools throughout the research process.

\section{Prompt Templates}
\label{app:prompts}
\subsection{EDGE: Evidence Detection via Grounding}

\begin{tcolorbox}[title=EDGE (Evidence Detection via Grounding), breakable]

Task - Ground Visual Evidence

You are given:

\begin{itemize}
\item A scientific or schematic diagram (e.g., flowchart, life cycle, bar chart, map, circuit)
\item A question about the diagram
\item The correct answer
\end{itemize}

Your task is to identify all visual regions required to verify the answer.

This includes:

\begin{itemize}
\item The visual region that directly depicts the answer
\item The textual label or annotation naming the answer
\item Any supporting visual evidence used for reasoning
\end{itemize}

Examples of supporting evidence include legends, arrows, connections, neighbouring regions, chart axes, or other visual structures used to interpret the diagram.

\textbf{What to Ground}

Include bounding boxes for:

\begin{itemize}
\item The answer region (visual depiction + label)
\item The question subject or anchor (e.g., a specific state, component, or object)
\item Any visual elements required for comparison, flow, or spatial reasoning
\item Supporting diagram features such as legends, axes, arrows, connectors, or colour encodings
\end{itemize}

Only include regions that a human would reasonably inspect to answer the question.

\textbf{Input}

Question ID: \{Question\_ID\}  
Question: \{Question\}  
Choices: \{Choices\}  
Answer: \{Answer\}  
Image metadata: \{height\}$\times$\{width\}

\textbf{Output Format (STRICT)}

\texttt{Answer: <ID1>/<x,y,w,h>, <ID2>/<x,y,w,h>, ...}

\texttt{Explanation: brief justification for each region}

Use only region identifiers and coordinates in the \texttt{Answer} field.

\textbf{Important}

\begin{itemize}
\item Do not return only the text label of the answer.
\item Also ground the visual region corresponding to the answer.
\item Do not output step-by-step reasoning.
\item Start your response directly with \texttt{Answer:}.
\end{itemize}

Example Output: \\

\texttt{Answer:R\_01/300,240,100,50, T\_01/305,250,80,40, L\_01/500,100,90,30, R\_02/250,240,90,50, R\_03/280,280,100,50, R\_04/330,220,90,40} \\ 

\texttt{Explanation:R\_01 is Mississippi, the correct answer. T\_01 is the label for Mississippi. L\_01 is the legend indicating the color coding of values. R\_02 to R\_04 are neighboring states used in comparison to determine the lowest value.}\\

\end{tcolorbox}

\subsection{SAGE: Select and Ground Evidence}

\paragraph{Stage 1: Evidence Selection}

\begin{tcolorbox}[title=SAGE Stage 1 (Select Visual Evidence), breakable]

Task - Identify Visual Evidence (Stage 1 of 2)

Given this diagram, a question, and its answer, list \textbf{every visual element} a human would inspect to verify the answer. \textbf{Do not predict bounding boxes yet.}

For each element, provide:
\begin{itemize}
\item A short identifier (e.g., B\_01, T\_01, R\_01)
\item A one-line description of the element
\end{itemize}

Be exhaustive. At minimum include:

\begin{enumerate}
\item \textbf{Answer Region(s)} - visual areas that directly depict the answer (bars, slices, states, nodes, lines, components, map regions, etc.). List each region separately.
\item \textbf{Labels and Text} - textual annotations, axis ticks, titles, or legend entries that name or quantify an answer region.
\item \textbf{Context and Supporting Evidence} - legends, color keys, axes, scale bars, arrows, connectors, or neighboring regions used for comparison or interpretation.
\end{enumerate}

Err on the side of including \textbf{more elements rather than fewer}.

\textbf{Input}

Question ID: \{Question\_ID\}  
Question: \{Question\}  
Choices: \{Choices\}  
Answer: \{Answer\}  
Image metadata: \{height\}$\times$\{width\}

\textbf{Output Format}

Start your response with:

\texttt{Elements: <ID1> - description, <ID2> - description, ...}

\end{tcolorbox}

\paragraph{Stage 2: Bounding Box Grounding}

\begin{tcolorbox}[title=SAGE Stage 2 (Ground Bounding Boxes), breakable]

Task - Bounding Box Grounding (Stage 2 of 2)

You previously identified the following visual elements required to answer a question about this diagram:

\textbf{Element List}

\{Stage1\_Elements\}

Provide a bounding box for \textbf{every element listed above}.

Each bounding box should be in pixel coordinates:

\begin{itemize}
\item $(x, y)$ = top-left corner
\item $(w, h)$ = width and height
\end{itemize}

Boxes should tightly enclose the corresponding visual element.

Do not skip any elements. If you identify additional answer-relevant elements that were missed in Stage 1, add them using new identifiers.

\textbf{Output Format}

\texttt{Answer: <ID1>/<x,y,w,h>, <ID2>/<x,y,w,h>, ...}

\texttt{Explanation: brief reason for each box}

Example:

\texttt{Answer: B\_01/120,45,200,80, T\_01/50,10,300,25}

Image metadata: \{height\}$\times$\{width\}

Start your response with \texttt{Answer:}

\end{tcolorbox}

\subsection{VERGE: Verify and Refine Grounded Evidence}

\begin{tcolorbox}[title=VERGE (Verify and Refine Grounding), breakable]

Task - Verify and Refine Grounding

You previously predicted bounding boxes for this diagram.  
Re-examine the image carefully and improve the grounding.

\textbf{Previous Prediction}

\{previous\_prediction\}

Perform both of the following steps:

\textbf{A. Fix Coordinates}

For each predicted bounding box:

\begin{itemize}
\item Verify that $(x,y)$ correctly marks the top-left corner.
\item Ensure $(w,h)$ tightly encloses the visual element.
\item Adjust coordinates if the box is misplaced, too large, or too small.
\end{itemize}

\textbf{B. Add Missing Evidence}

Add any visual evidence that supports verifying the answer, including:

\begin{itemize}
\item Answer regions
\item Labels and textual annotations
\item Legends or colour keys
\item Axes or scale bars
\item Arrows and connectors
\item Neighbouring regions used for comparison
\end{itemize}

Err on the side of including \textbf{more boxes rather than fewer}.

\textbf{Rules}

\begin{itemize}
\item Never remove an existing bounding box.
\item Only adjust coordinates or add new boxes.
\item Use identifiers such as B\_01, T\_01, R\_01 for new regions.
\item Coordinates are specified in pixels for image size
\{height\}$\times$\{width\}.
\end{itemize}

\textbf{Output Format}

\texttt{Answer: <ID1>/<x,y,w,h>, <ID2>/<x,y,w,h>, ...}

\texttt{Explanation: describe coordinate fixes and newly added evidence.}

Return the \textbf{complete set of boxes} (all original boxes plus any new ones).  
Start your response with \texttt{Answer:}.

\end{tcolorbox}

\section{Experimentation Details}
\label{app:exp_details}

This appendix provides the full implementation details for the experimental evaluation described in Section~\ref{sec:exp}. We describe the model configurations, inference setup, image preprocessing pipeline, infrastructure, and the detailed definition of evaluation metrics.


We evaluate all models using the \benchmark{} test split. For each sample, the input consists of a diagram image, a question, and the verified answer. The model receives one of the prompting strategies described in Section~\ref{sec:modeling}. The prompt instructs the model to predict visual evidence regions that justify the answer.

All models produce outputs in a standardized bounding-box format:

\begin{center}
\texttt{ID/x,y,w,h}
\end{center}

where $(x,y)$ denotes the top-left corner of the box and $(w,h)$ denotes the width and height in pixel coordinates. We extract predicted boxes using regular-expression parsing and discard malformed outputs.

We apply a consistent decoding configuration across models whenever the inference interface exposes the relevant parameters. Specifically, we set the temperature to 0.2 to reduce stochastic variation and use top-$p$ sampling with $p=0.7$. We allow a maximum output length of 2048 tokens for EDGE prompting methods and 1024 tokens per stage for two-stage and self-correction prompting strategies. For locally hosted models, we apply a repetition penalty of 1.1 to reduce duplicated bounding-box outputs. To improve robustness during large-scale inference, we implement several reliability mechanisms. For AWS Bedrock inference requests, the system automatically retries failed calls up to five times. For locally hosted GPU inference, if a batch triggers an out-of-memory failure, we rerun the failed samples individually to ensure the evaluation proceeds without data loss.

\subsection{Model Configuration}

We evaluate eight recent vision-language models spanning both proprietary API systems and open-weight checkpoints. Our evaluation includes Claude-Opus-4.6 (\nolinkurl{us.anthropic.claude-opus-4-6-v1}), Claude-Sonnet-4.6 (\nolinkurl{us.anthropic.claude-sonnet-4-6}), and, all accessed through AWS Bedrock. We also evaluate Gemini-3-Pro (\nolinkurl{gemini-3-pro-preview}) through Google Vertex AI Batch inference and Llama-4-Maverick-17B-IT (\nolinkurl{us.meta.llama4-maverick-17b-instruct-v1:0}) through AWS Bedrock.

For open-weight models, we run Qwen3.5-35B-A3B (\nolinkurl{Qwen/Qwen3.5-35B-A3B}), Gemma3-27B-IT (\nolinkurl{google/gemma-3-27b-it}), and InternVL3.5-38B (\nolinkurl{InternVL3.5-38B}) locally using HuggingFace-based inference pipelines. Qwen3.5-35B-A3B and InternVL3.5-38B require larger memory footprints and therefore run across two H200 GPUs using automatic device sharding, while Gemma3-27B-IT runs on a single GPU. These deployment configurations ensure stable inference while accommodating the memory requirements of large multimodal models.

We distribute the evaluation across both cloud-based APIs and local GPU infrastructure to maintain reliability and redundancy during large-scale inference. API-based models run entirely through managed inference platforms such as AWS Bedrock and Google Vertex AI, while open-weight models execute on local GPUs. This hybrid setup allows us to handle infrastructure failures gracefully, isolate model-specific runtime requirements, and ensure that evaluation remains consistent across models with different deployment constraints. We evaluate all models using their released checkpoints and do not apply any additional fine-tuning.

\subsection{Metrics Details}

Before computing any metric, we normalize all predicted and ground-truth box coordinates by image width and height so that every coordinate lies in the interval $[0,1]$. This normalization ensures comparability across images with different resolutions.

\paragraph{Max Pairwise IoU (MP$_{\mathrm{IoU}}$)}

Max Pairwise IoU measures the best-case localization quality of a model. For each sample, we compute the intersection-over-union (IoU) between every predicted box and every ground-truth box:

\[
IoU = \frac{\text{Area}(B_p \cap B_{gt})}{\text{Area}(B_p \cup B_{gt})}
\]

where $B_p$ denotes a predicted box and $B_{gt}$ denotes a ground-truth box.

We retain the maximum IoU across all box pairs for a given sample and then average this value across the dataset. This metric evaluates whether the model predicts at least one region that aligns closely with the ground-truth evidence.

\paragraph{Grounding IoU (G$_{\mathrm{IoU}}$)}

Grounding IoU evaluates how well the entire predicted evidence set aligns with the full ground-truth evidence set. Instead of comparing boxes individually, we compute the union of all predicted boxes and the union of all ground-truth boxes and then measure the IoU between these two merged regions.

\[
IoU_{group} =
\frac{\text{Area}(\cup B_p \cap \cup B_{gt})}
{\text{Area}(\cup B_p \cup \cup B_{gt})}
\]

This metric penalizes excessive predictions because unnecessary boxes increase the union of predicted regions and reduce overlap quality. We compute region unions using polygon operations implemented in the Shapely geometry library, following evaluation strategies used in grounding benchmarks such as GRIT.

\paragraph{Thresholded Metrics.}

We also compute thresholded hit rates for both metrics. For a threshold $\tau$, MaxIoU@$ \tau $ marks a sample as correct when its maximum pairwise IoU exceeds $\tau$. Similarly, GroundingIoU@$ \tau $ marks a sample as correct when the merged-region IoU exceeds $\tau$. These thresholds allow us to evaluate localization quality under progressively stricter alignment requirements.

\paragraph{Precision, Recall, and F1.}

Finally, we evaluate box-level detection quality using precision, recall, and F1. At threshold $\tau$, a predicted box counts as a true positive if it overlaps with at least one ground-truth box with IoU $\geq \tau$. A predicted box counts as a false positive if it fails to match any ground-truth box. A ground-truth box counts as a false negative if no predicted box overlaps it at the threshold.

For each sample, recall measures the fraction of ground-truth boxes covered by predictions, and precision measures the fraction of predicted boxes that correspond to valid evidence regions. After computing recall and precision for each sample, we average these values across the dataset and compute F1 as

\[
F1 =
\frac{2 \cdot \overline{P} \cdot \overline{R}}
{\overline{P} + \overline{R}}.
\]

When a sample contains no predicted boxes or no ground-truth boxes, we assign zero to precision, recall, and F1 for that sample.

\section{Annotation Instructions}
\label{appendix:annotation_instructions}

We provided annotators with a standardized protocol to ensure consistency in evidence selection across datasets and diagram types. The protocol defined the annotation objective, the review procedure, and the arbitration policy used to resolve disagreements.

\subsection{Annotation Objective}

Annotators identified the minimal set of visual regions required to verify the answer to each question. They did not restrict annotations to the final answer region. Instead, they selected all visual regions necessary to support the reasoning process, including intermediate evidence when the diagram required multi-step interpretation.

\subsection{Review Procedure}

Annotators followed three steps for each instance:

\begin{enumerate}
\setlength\itemsep{0em}
\item They read the question carefully and identified all referenced visual elements in the diagram.
\item They used the verified answer, and answer choices when available, to determine which regions were necessary to justify the answer.
\item They flagged ambiguous, inconsistent, or visually unanswerable items for arbitration rather than guessing.
\end{enumerate}

\subsection{Annotation Policy}

We used a multi-annotator verification procedure to ensure reliability.

\begin{enumerate}
\setlength\itemsep{0em}
\item Two annotators independently reviewed each instance (six annotators participated overall).

\item A senior annotator resolved disagreements using predefined arbitration guidelines. 
\item We retained only consensus or majority-agreed evidence annotations in the final dataset.
\end{enumerate}

\subsection{Instructional Support}

We provided dataset-specific examples covering scientific diagrams, charts, maps, circuits, and infographics to illustrate reasoning-level evidence selection. We include representative qualitative examples in Appendix~\ref{appendix:qualitative}.

\section{Inter-Annotator Agreement Details}
\label{appendix:iaa}

\subsection{Evaluation Scope}

We evaluate annotation reliability for the selection-based proposal workflow only. In this setting, annotators select evidence from a pre-existing candidate region set. We exclude newly drawn regions from this analysis so that the agreement measure reflects evidence selection quality rather than variation introduced by box creation.

\subsection{Evaluation Criteria}

We evaluate agreement using two complementary criteria.

\paragraph{Complete Visual Reasoning (CVR).}
Annotators mark an instance as correct only when the selected regions cover the full visual reasoning chain required to derive the answer. If any required intermediate evidence region is missing, they mark the instance as incorrect.

\paragraph{Core Evidence Alignment (CEA).}
Annotators mark an instance as correct when the selected regions capture the primary evidence needed to answer the question, even if minor intermediate components are omitted. This criterion measures agreement on essential grounding rather than exhaustive coverage.

These criteria distinguish between strict reasoning-chain completeness and agreement on core evidence. Together, they provide a more informative view of annotation reliability for evidence-grounded reasoning than a single binary criterion.

\paragraph{Output format for reproducibility.}
Our evaluation suite expects a strict, machine-parseable output format with an \texttt{Answer:} line containing comma-separated \texttt{ID/<x,y,w,h>} entries, followed by an \texttt{Explanation:} line for optional free-form justification. This makes it straightforward to evaluate black-box APIs while keeping the task definition model-agnostic.


\section{Qualitative Analysis of Evidence Grounding}
\label{appendix:qualitative}

We present qualitative grounding results across all eight models and three 
prompting strategies to complement the quantitative analysis in 
Section~\ref{sec:results}. These examples illustrate characteristic success 
and failure modes across both model families and prompt designs on 
representative instances from the six domains.

\begin{figure*}[t]
\centering
\includegraphics[width=1\linewidth]{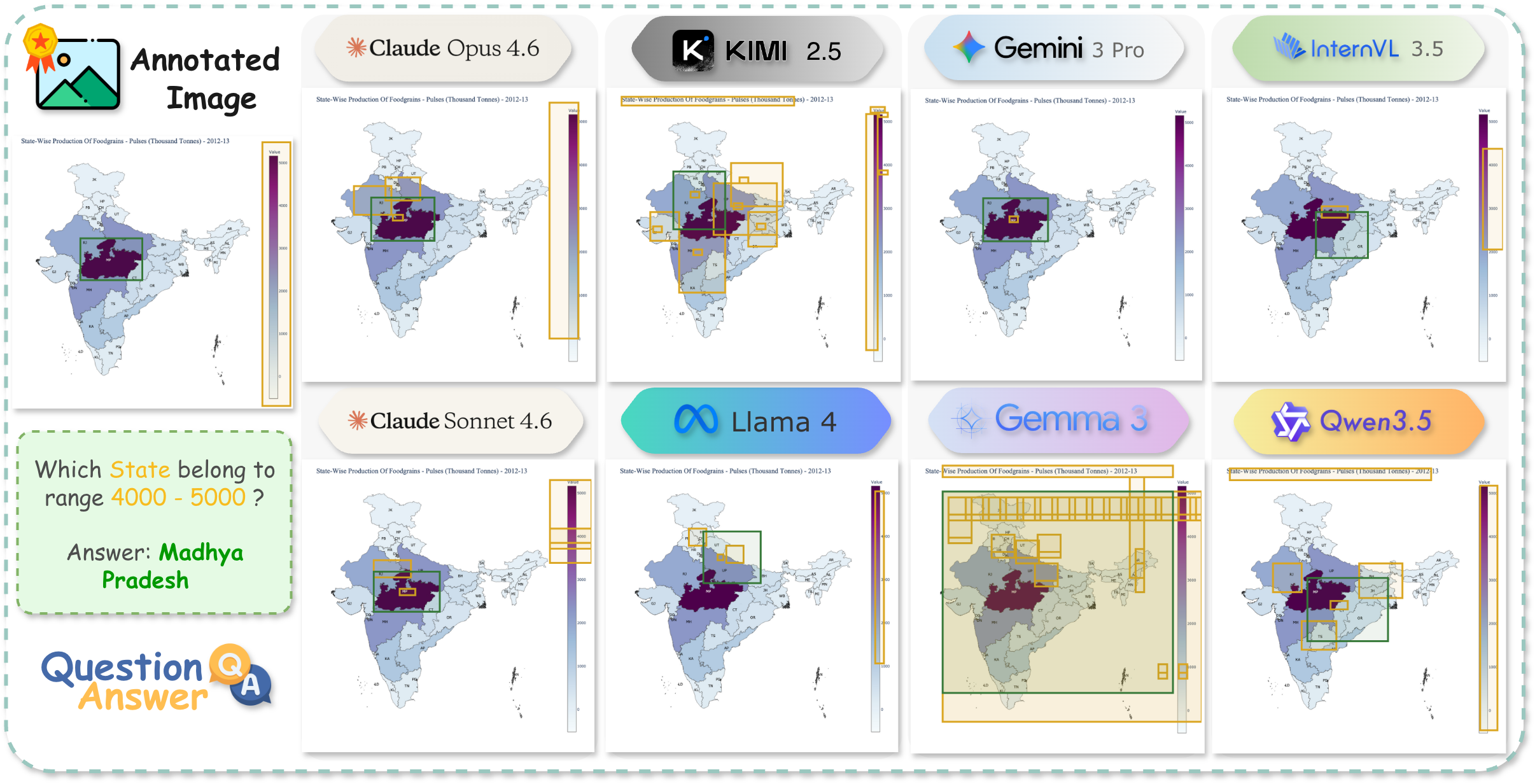}
\caption{Model-wise success case on a MapWise instance. The 
annotated image (left) shows the gold evidence: a green box identifying 
Madhya Pradesh as the state falling in the production range 4000-5000 
thousand tonnes, and an orange box marking the corresponding legend 
segment. Each model panel shows predicted bounding boxes (green for 
answer regions, orange for supporting evidence). Closed-source models 
produce compact, well-localized predictions, while open-weight models 
exhibit varying degrees of over-prediction and spatial imprecision.}
\label{fig:model_success}
\end{figure*}



\begin{figure*}[t]
\centering
\includegraphics[width=\linewidth]{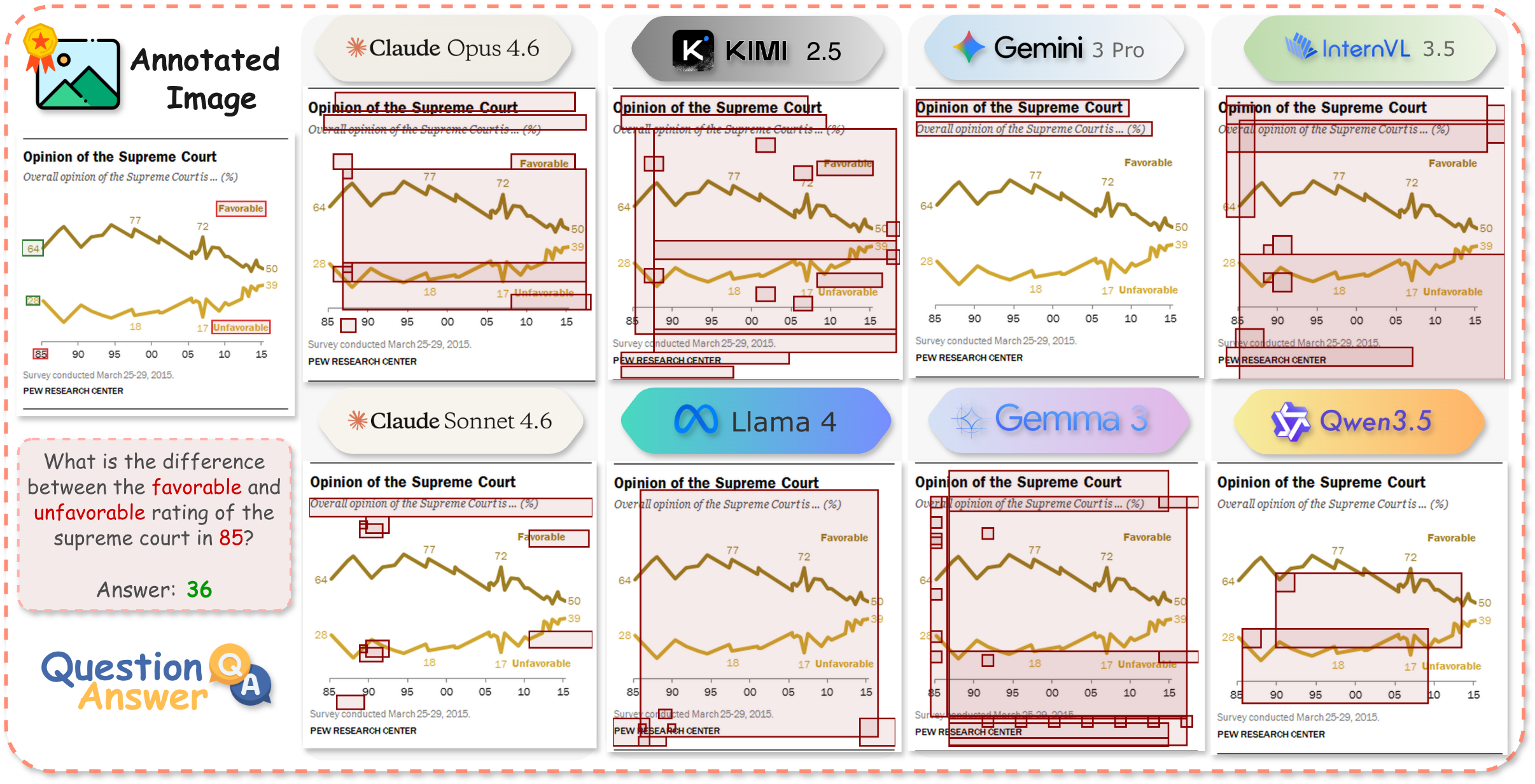}
\caption{Model-wise failure case on a ChartQA instance. The 
annotated image (left) shows the gold evidence: green boxes marking the 
favorable (64) and unfavorable (28) data points at year 1985, whose 
difference yields the answer of 36. All eight models fail to precisely 
localize the two target data points simultaneously, instead grounding 
chart titles, subtitles, full trend lines, or axis regions that do not 
constitute the minimal reasoning evidence.}
\label{fig:model_failure}
\end{figure*}



\begin{figure*}[t]
\centering
\includegraphics[width=\linewidth]{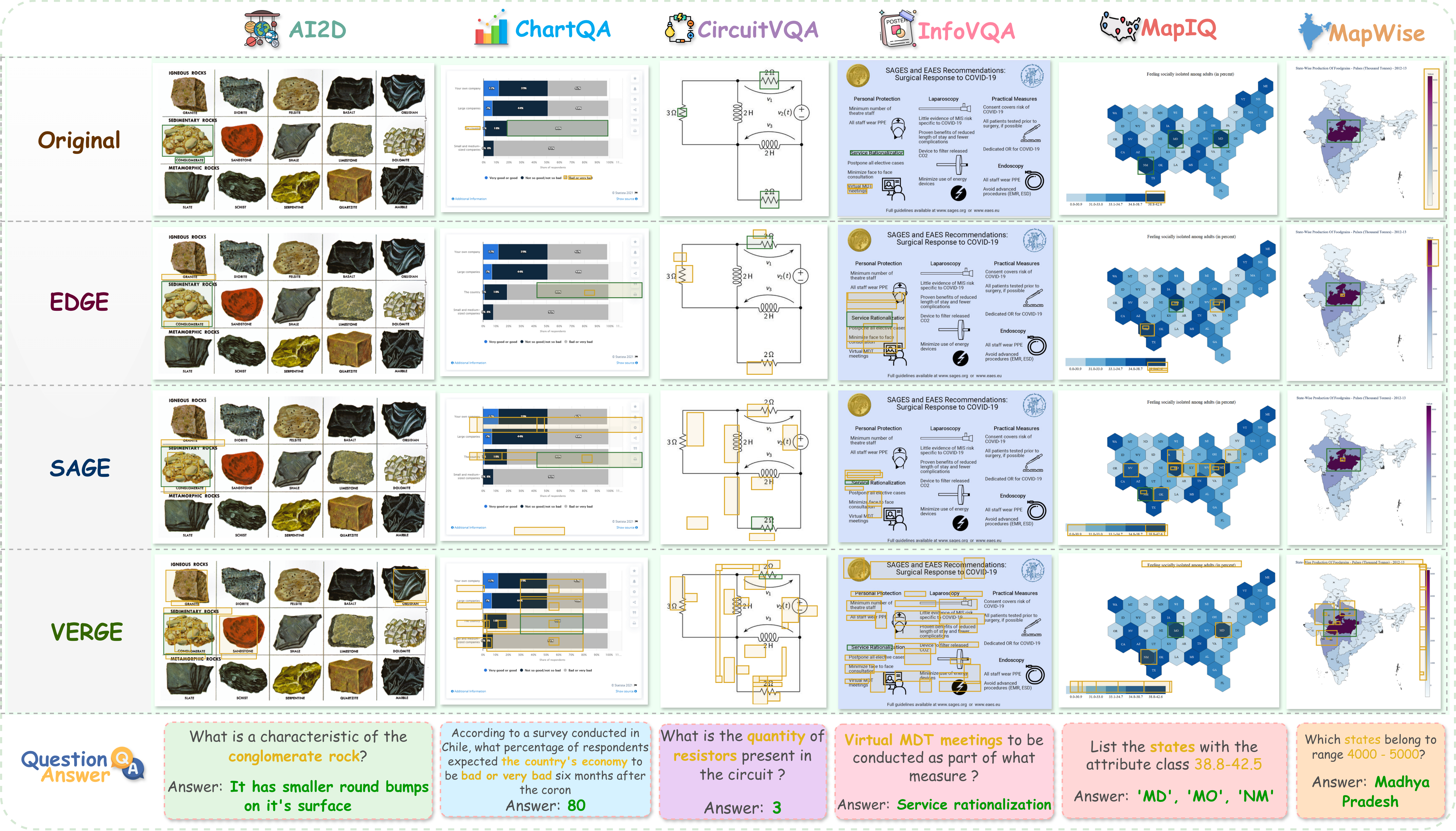}
\caption{Prompt-wise success cases across all six \benchmark 
domains. Each column shows one diagram domain; rows show the ground-truth 
annotations (Original), and predictions under EDGE, SAGE, and VERGE. 
Across domains, all three prompting strategies successfully localize the 
core evidence regions, with VERGE producing the most complete coverage in 
most cases. Questions and answers are shown at the bottom of each column.}
\label{fig:prompt_success}
\end{figure*}



\begin{figure*}[t]
\centering
\includegraphics[width=\linewidth]{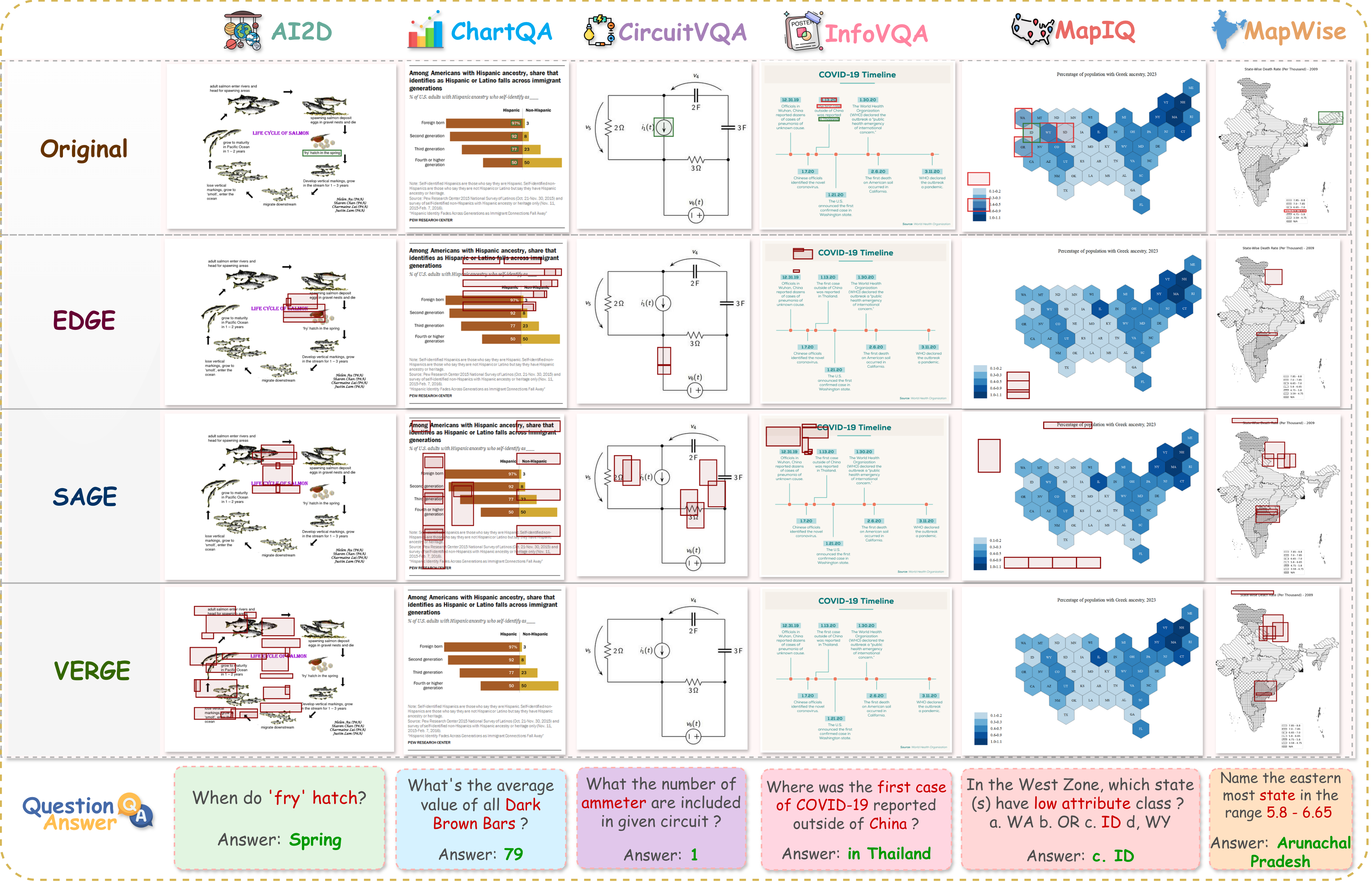}
\caption{Prompt-wise failure cases across all six \benchmark 
domains. Each column shows one diagram domain; rows show the ground-truth 
annotations (Original), and predictions under the three prompting techniques: EDGE, SAGE, and VERGE. Across domains, all three prompting strategies fail to recover the complete gold evidence set, with characteristic failure modes that persist regardless of prompt design: over-prediction in structurally dense 
diagrams, missing evidence in multi-hop reasoning instances, and 
imprecise localization in spatially ambiguous regions.}
\label{fig:prompt_failure}
\end{figure*}

\begin{table*}[t]
\caption{Grounding performance of additional models across prompting strategies and diagram domains. 
This table complements Table~\ref{tab:prompt_dataset_model} by extending the analysis to open-weight and smaller models. 
Results show that (i) prompting effects are highly model-dependent and inconsistent across domains, and (ii) weaker models often exhibit non-monotonic behavior across EDGE, SAGE, and VERGE, with occasional gains in localization (MP$_{\mathrm{IoU}}$) that do not translate into improved evidence coverage (G$_{\mathrm{IoU}}$, F1). 
Overall, the results highlight that prompting alone is insufficient to close the grounding gap for lower-capacity models.}
\centering
\small
\setlength{\tabcolsep}{3pt}
\renewcommand{\arraystretch}{1.3}

\resizebox{\textwidth}{!}{
\begin{tabular}{l|ccc|ccc|ccc|ccc|ccc|ccc}

\toprule
& \multicolumn{3}{c|}{\textbf{ChartQA}}
& \multicolumn{3}{c|}{\textbf{Circuit-VQA}}
& \multicolumn{3}{c|}{\textbf{InfoVQA}}
& \multicolumn{3}{c|}{\textbf{MapIQ}}
& \multicolumn{3}{c|}{\textbf{MapWise}}
& \multicolumn{3}{c|}{\textbf{AI2D}} \\

\cmidrule(lr){2-4}
\cmidrule(lr){5-7}
\cmidrule(lr){8-10}
\cmidrule(lr){11-13}
\cmidrule(lr){14-16}
\cmidrule(lr){17-19}

\textbf{Prompt}
& \makecell{\textbf{MP$_{\mathrm{IoU}}$}} & \makecell{\textbf{G$_{\mathrm{IoU}}$}} & \textbf{F1}
& \makecell{\textbf{MP$_{\mathrm{IoU}}$}} & \makecell{\textbf{G$_{\mathrm{IoU}}$}} & \textbf{F1}
& \makecell{\textbf{MP$_{\mathrm{IoU}}$}} & \makecell{\textbf{G$_{\mathrm{IoU}}$}} & \textbf{F1}
& \makecell{\textbf{MP$_{\mathrm{IoU}}$}} & \makecell{\textbf{G$_{\mathrm{IoU}}$}} & \textbf{F1}
& \makecell{\textbf{MP$_{\mathrm{IoU}}$}} & \makecell{\textbf{G$_{\mathrm{IoU}}$}} & \textbf{F1}
& \makecell{\textbf{MP$_{\mathrm{IoU}}$}} & \makecell{\textbf{G$_{\mathrm{IoU}}$}} & \textbf{F1}\\

\hline
\multicolumn{19}{>{\columncolor{lightestgray}}c}{\textbf{Claude Sonnet 4.6}} \\
\hline
\textbf{EDGE}  & 15.0 & 10.1 & 3.4 & 4.3 & 2.3 & 0.1 & 4.0 & 2.4 & 0.1 & 47.2 & 26.5 & 16.4 & 25.0 & 11.1 & 5.0 & 29.7 & 21.3 & 7.4 \\
\textbf{SAGE}  & 5.6 & 2.4 & 0.5 & 2.0 & 0.9 & 0.1 & 0.7 & 0.3 & 0.0 & 20.3 & 10.7 & 7.7 & 16.5 & 6.5 & 2.8 & 7.1 & 5.0 & 1.5 \\
\textbf{VERGE} & 20.5 & 10.6 & 3.1 & 6.9 & 3.3 & 0.2 & 5.0 & 2.1 & 0.1 & 54.2 & 26.8 & 17.3 & 30.1 & 11.6 & 4.8 & 34.6 & 21.3 & 7.5 \\

\hline
\multicolumn{19}{>{\columncolor{lightestgray}}c}{\textbf{Llama4-Maverick-17B-IT}} \\
\hline
\textbf{EDGE}  & 11.7 & 8.1 & 1.3 & 4.7 & 2.9 & 0.2 & 3.2 & 2.8 & 0.1 & 24.7 & 10.8 & 1.8 & 18.3 & 8.6 & 2.3 & 19.9 & 14.8 & 4.1 \\
\textbf{SAGE}  & 0.0 & 0.0 & 0.0 & 1.1 & 0.3 & 0.1 & 0.1 & 0.1 & 0.0 & 0.2 & 0.0 & 0.0 & 0.2 & 0.2 & 0.0 & 0.6 & 0.4 & 0.0 \\
\textbf{VERGE} & 11.4 & 6.6 & 0.6 & 4.9 & 2.3 & 0.4 & 2.4 & 1.5 & 0.0 & 22.1 & 7.7 & 1.3 & 14.8 & 6.7 & 1.0 & 19.0 & 13.4 & 3.0 \\

\hline
\multicolumn{19}{>{\columncolor{lightestgray}}c}{\textbf{Qwen3.5-35B-A3B}} \\
\hline
\textbf{EDGE}  & 10.7 & 6.6 & 1.2 & 4.2 & 2.4 & 0.3 & 1.9 & 1.7 & 0.1 & 20.3 & 3.7 & 1.1 & 11.2 & 4.2 & 1.2 & 12.6 & 8.7 & 1.3 \\
\textbf{SAGE}  & 11.5 & 4.2 & 0.3 & 3.9 & 1.8 & 0.2 & 2.3 & 0.9 & 0.0 & 12.6 & 5.1 & 0.4 & 10.5 & 3.5 & 0.6 & 14.1 & 7.3 & 1.0 \\
\textbf{VERGE} & 15.6 & 6.3 & 0.9 & 8.0 & 2.0 & 0.2 & 3.6 & 1.2 & 0.0 & 22.0 & 5.3 & 1.0 & 15.2 & 5.1 & 0.7 & 22.1 & 11.5 & 1.3 \\

\hline
\multicolumn{19}{>{\columncolor{lightestgray}}c}{\textbf{Gemma3-27B-IT}} \\
\hline
\textbf{EDGE}  & 3.8 & 2.5 & 0.1 & 2.7 & 1.2 & 0.2 & 1.2 & 0.6 & 0.0 & 9.2 & 2.4 & 0.3 & 9.3 & 4.4 & 0.7 & 12.7 & 8.5 & 1.8 \\
\textbf{SAGE}  & 11.8 & 5.3 & 0.4 & 9.5 & 2.0 & 0.2 & 7.4 & 1.5 & 0.1 & 29.0 & 7.1 & 1.3 & 22.5 & 4.9 & 0.6 & 27.2 & 16.5 & 2.2 \\
\textbf{VERGE} & 8.4 & 4.4 & 0.1 & 7.7 & 1.7 & 0.2 & 4.0 & 1.1 & 0.1 & 18.2 & 3.9 & 0.6 & 13.9 & 6.1 & 0.9 & 17.1 & 12.1 & 2.0 \\

\bottomrule
\end{tabular}
}

\label{tab:prompt_dataset_model_others}

\end{table*}


\begin{table}[t]
\caption{
Grounding performance at IoU threshold $\tau=0.3$, measuring coarse localization ability. 
At this threshold, predictions are considered correct if they approximately overlap the target region, even without tight alignment. 
Results show that most models achieve substantially higher MP$_{\mathrm{IoU}}$ and G$_{\mathrm{IoU}}$ compared to stricter thresholds, indicating that models can often identify roughly relevant regions. 
However, the gap between MP$_{\mathrm{IoU}}$ and F1 persists across all datasets, suggesting that coarse localization does not imply complete evidence grounding.
}
\centering
\scriptsize
\setlength{\tabcolsep}{0.5pt}
\renewcommand{\arraystretch}{1.1}

\begin{tabular}{l|ccc|ccc|ccc|ccc|ccc|ccc}
\toprule

& \multicolumn{3}{c|}{\textbf{ChartQA}} 
& \multicolumn{3}{c|}{\textbf{CircuitVQA}}
& \multicolumn{3}{c|}{\textbf{InfographicVQA}}
& \multicolumn{3}{c|}{\textbf{MapIQ}}
& \multicolumn{3}{c|}{\textbf{Mapwise}}
& \multicolumn{3}{c}{\textbf{AI2D}} \\

\cmidrule(lr){2-4} \cmidrule(lr){5-7} \cmidrule(lr){8-10}
\cmidrule(lr){11-13} \cmidrule(lr){14-16} \cmidrule(lr){17-19}

Model
& \makecell{\textbf{MP$_{\mathrm{IoU} }$}} & \makecell{\textbf{G$_{\mathrm{IoU}}$}} & \textbf{F1}
& \makecell{\textbf{MP$_{\mathrm{IoU}}$}} & \makecell{\textbf{G$_{\mathrm{IoU}}$}} & \textbf{F1}
& \makecell{\textbf{MP$_{\mathrm{IoU}}$}} & \makecell{\textbf{G$_{\mathrm{IoU}}$}} & \textbf{F1}
& \makecell{\textbf{MP$_{\mathrm{IoU}}$}} & \makecell{\textbf{G$_{\mathrm{IoU}}$}} & \textbf{F1}
& \makecell{\textbf{MP$_{\mathrm{IoU}}$}} & \makecell{\textbf{G$_{\mathrm{IoU}}$}} & \textbf{F1}
& \makecell{\textbf{MP$_{\mathrm{IoU}}$}} & \makecell{\textbf{G$_{\mathrm{IoU}}$}} & \textbf{F1}\\

\hline
\multicolumn{19}{>{\columncolor{lightestgray}}c}{\textbf{EDGE}} \\
\hline

Claude Opus 4.6  & 41.4 & 15.2 & 11.8 & 6.9 & 1.5 & 3.0 & 10.5 & 1.9 & 2.7 & 88.1 & 49.3 & 35.2 & 28.5 & 4.1 & 8.5 & 53.2 & 31.3 & 19.6 \\
Claude Sonnet 4.6 & 18.0 & 10.8 & 6.1 & 3.7 & 0.7 & 1.2 & 2.4 & 1.0 & 0.8 & 76.7 & 38.5 & 30.5 & 40.6 & 7.7 & 16.2 & 45.3 & 29.6 & 18.6 \\
Kimi K2.5         & 46.0 & 27.5 & 16.1 & 8.2 & 3.0 & 3.0 & 11.7 & 1.7 & 4.4 & 62.6 & 10.6 & 15.1 & 38.0 & 4.4 & 12.0 & 46.2 & 22.2 & 18.2 \\
Gemini 3 Pro       & 22.4 & 10.8 & 9.0  & 13.4 & 4.7 & 7.2 & 11.9 & 6.3 & 5.2 & 19.2 & 5.4 & 4.3 & 27.2 & 14.9 & 13.1 & 40.8 & 22.4 & 19.0 \\
Llama4-Maverick-17B-IT      & 16.5 & 10.5 & 5.3  & 3.5 & 1.0 & 1.4 & 1.0 & 1.0 & 0.3 & 40.1 & 7.0 & 9.4 & 30.1 & 4.6 & 11.2 & 30.0 & 16.6 & 14.8 \\
Qwen3.5-35B-A3B      & 12.3 & 5.9 & 3.8   & 3.5 & 0.7 & 1.4 & 0.5 & 0.5 & 0.2 & 38.2 & 0.5 & 5.7 & 15.9 & 1.5 & 4.9 & 16.1 & 8.5 & 6.6 \\
Gemma3-27B-IT       & 2.6 & 1.8 & 1.1    & 2.2 & 0.2 & 0.8 & 0.2 & 0.0 & 0.1 & 10.6 & 0.3 & 1.9 & 12.6 & 1.5 & 3.9 & 14.5 & 7.2 & 6.4 \\
InternVL 3.5-38B     & 6.2 & 2.8 & 1.7    & 1.2 & 0.7 & 0.7 & 0.7 & 0.0 & 0.1 & 11.4 & 0.3 & 1.9 & 5.4 & 0.8 & 1.9 & 10.4 & 5.2 & 4.7 \\

\hline
\multicolumn{19}{>{\columncolor{lightestgray}}c}{\textbf{SAGE}} \\
\hline

Claude Opus 4.6  & 14.7 & 8.7 & 3.3 & 12.1 & 1.2 & 2.0 & 18.5 & 1.5 & 3.8 & 89.4 & 32.5 & 30.6 & 35.0 & 2.8 & 6.0 & 68.3 & 31.1 & 18.2 \\
Claude Sonnet 4.6 & 7.7 & 1.8 & 1.4 & 1.7 & 0.2 & 0.5 & 0.7 & 0.0 & 0.2 & 30.9 & 16.8 & 12.0 & 27.0 & 4.1 & 7.8 & 11.8 & 7.0 & 3.8 \\
Kimi K2.5          & 48.6 & 18.0 & 10.3 & 15.3 & 0.0 & 1.2 & 12.9 & 0.7 & 2.0 & 83.5 & 17.3 & 17.5 & 68.1 & 7.5 & 12.8 & 66.3 & 28.8 & 15.3 \\
Gemini 3 Pro        & 31.6 & 19.8 & 12.6 & 11.9 & 4.0 & 7.2 & 16.3 & 10.5 & 6.7 & 11.1 & 2.2 & 3.3 & 12.6 & 5.4 & 6.4 & 26.5 & 12.0 & 13.7 \\
Llama4-Maverick-17B-IT        & 0.0 & 0.0 & 0.0 & 0.7 & 0.0 & 0.2 & 0.0 & 0.0 & 0.0 & 0.3 & 0.0 & 0.0 & 0.5 & 0.5 & 0.2 & 0.8 & 0.6 & 0.2 \\
Qwen3.5-35B-A3B       & 15.4 & 2.3 & 3.1 & 3.7 & 1.0 & 1.5 & 1.5 & 0.0 & 0.3 & 16.0 & 1.4 & 2.6 & 13.1 & 1.3 & 3.1 & 16.4 & 5.6 & 5.7 \\
Gemma3-27B-IT      & 11.1 & 3.3 & 1.5 & 8.9 & 0.0 & 0.7 & 7.5 & 1.2 & 0.5 & 48.5 & 1.4 & 5.8 & 36.0 & 0.0 & 3.4 & 42.9 & 21.1 & 7.5 \\
InternVL 3.5-38B       & 13.1 & 1.3 & 2.8 & 7.9 & 0.5 & 1.5 & 3.6 & 1.2 & 0.7 & 3.3 & 0.0 & 0.3 & 6.4 & 0.3 & 1.4 & 29.0 & 11.8 & 7.0 \\

\hline
\multicolumn{19}{>{\columncolor{lightestgray}}c}{\textbf{VERGE}} \\
\hline

Claude Opus 4.6  & 55.8 & 18.3 & 11.5 & 11.4 & 1.0 & 2.8 & 14.6 & 1.5 & 3.3 & 96.2 & 44.2 & 33.8 & 38.6 & 3.9 & 7.0 & 66.5 & 27.7 & 18.0 \\
Claude Sonnet 4.6 & 25.4 & 11.3 & 6.6 & 4.7 & 1.0 & 1.6 & 4.9 & 0.5 & 1.0 & 87.8 & 39.6 & 33.2 & 48.8 & 8.0 & 13.4 & 56.5 & 26.3 & 18.1 \\
Kimi K2.5         & 49.1 & 15.2 & 10.9 & 10.9 & 0.2 & 1.8 & 11.7 & 0.2 & 2.7 & 74.5 & 11.9 & 13.5 & 45.0 & 2.6 & 9.5 & 53.0 & 19.0 & 11.8 \\
Gemini 3 Pro       & 12.9 & 5.7 & 5.0 & 11.9 & 5.2 & 7.3 & 6.6 & 3.6 & 2.8 & 8.9 & 3.5 & 2.3 & 22.1 & 10.3 & 11.3 & 27.1 & 15.7 & 14.6 \\
Llama4-Maverick-17B-IT       & 12.9 & 5.4 & 2.9 & 4.5 & 1.2 & 1.4 & 1.5 & 0.2 & 0.5 & 37.4 & 4.3 & 6.7 & 21.3 & 2.3 & 5.8 & 26.9 & 13.9 & 8.9 \\
Qwen3.5-35B-A3B      & 18.8 & 4.1 & 3.8 & 6.7 & 0.2 & 1.1 & 1.5 & 0.0 & 0.1 & 39.3 & 0.0 & 5.6 & 21.6 & 1.0 & 3.2 & 31.3 & 10.6 & 6.4 \\
Gemma3-27B-IT       & 7.7 & 2.1 & 1.2 & 6.4 & 0.2 & 0.8 & 5.1 & 0.2 & 0.7 & 29.5 & 0.3 & 4.4 & 18.8 & 1.8 & 4.2 & 22.2 & 12.4 & 5.9 \\
InternVL 3.5-38B      & 10.3 & 2.6 & 2.3 & 4.2 & 0.0 & 1.2 & 0.5 & 0.2 & 0.2 & 11.1 & 0.0 & 2.0 & 5.1 & 0.5 & 1.7 & 20.5 & 8.9 & 6.2 \\

\bottomrule
\end{tabular}

\label{tab:threshold_table_0.3}
\end{table}

\begin{table*}[t]
\caption{
Grounding performance at IoU threshold $\tau=0.5$, requiring 50\% alignment between predicted and ground-truth regions. 
Compared to Table~\ref{tab:threshold_table_0.3}, performance drops sharply across all models and domains, especially in G$_{\mathrm{IoU}}$ and F1. 
This degradation highlights that many predictions that appear correct under coarse localization fail to precisely capture the intended evidence regions. 
The results emphasize that reasoning-level grounding requires not only identifying relevant regions but also accurately localizing all supporting evidence.
}
\centering
\scriptsize
\setlength{\tabcolsep}{0.5pt}
\renewcommand{\arraystretch}{1.1}

\begin{tabular}{l|ccc|ccc|ccc|ccc|ccc|ccc}
\toprule

& \multicolumn{3}{c|}{\textbf{ChartQA}} 
& \multicolumn{3}{c|}{\textbf{CircuitVQA}}
& \multicolumn{3}{c|}{\textbf{InfographicVQA}}
& \multicolumn{3}{c|}{\textbf{MapIQ}}
& \multicolumn{3}{c|}{\textbf{Mapwise}}
& \multicolumn{3}{c}{\textbf{AI2D}} \\

\cmidrule(lr){2-4} \cmidrule(lr){5-7} \cmidrule(lr){8-10}
\cmidrule(lr){11-13} \cmidrule(lr){14-16} \cmidrule(lr){17-19}

Model
& \makecell{\textbf{MP$_{\mathrm{IoU} }$}} & \makecell{\textbf{G$_{\mathrm{IoU}}$}} & \textbf{F1}
& \makecell{\textbf{MP$_{\mathrm{IoU}}$}} & \makecell{\textbf{G$_{\mathrm{IoU}}$}} & \textbf{F1}
& \makecell{\textbf{MP$_{\mathrm{IoU}}$}} & \makecell{\textbf{G$_{\mathrm{IoU}}$}} & \textbf{F1}
& \makecell{\textbf{MP$_{\mathrm{IoU}}$}} & \makecell{\textbf{G$_{\mathrm{IoU}}$}} & \textbf{F1}
& \makecell{\textbf{MP$_{\mathrm{IoU}}$}} & \makecell{\textbf{G$_{\mathrm{IoU}}$}} & \textbf{F1}
& \makecell{\textbf{MP$_{\mathrm{IoU}}$}} & \makecell{\textbf{G$_{\mathrm{IoU}}$}} & \textbf{F1}\\

\hline
\multicolumn{19}{>{\columncolor{lightestgray}}c}{\textbf{EDGE}} \\
\hline
Claude Opus 4.6
& 19.3 & 4.1 & 4.9 & 0.7 & 0.0 & 0.2 & 2.2 & 0.0 & 0.4 & 56.6 & 14.1 & 13.9 & 7.2 & 0.3 & 1.8 & 24.0 & 10.1 & 7.8 \\

Claude Sonnet 4.6
& 9.0 & 6.4 & 3.4 & 0.5 & 0.0 & 0.1 & 0.2 & 0.0 & 0.1 & 50.9 & 10.6 & 16.4 & 16.7 & 1.3 & 5.0 & 19.9 & 10.4 & 7.4 \\

Kimi K2.5     
& 30.6 & 11.1 & 9.7 & 1.2 & 0.2 & 0.3 & 3.2 & 0.2 & 1.2 & 24.1 & 0.3 & 2.7 & 9.5 & 0.0 & 2.0 & 20.9 & 6.4 & 7.3 \\

Gemini 3 Pro       
& 10.3 & 3.9 & 4.3 & 5.0 & 1.2 & 2.5 & 3.9 & 1.7 & 1.7 & 4.3 & 0.5 & 1.1 & 15.9 & 5.4 & 6.9 & 25.7 & 10.8 & 12.1 \\

Llama4-Maverick-17B-IT       
& 3.6 & 1.0 & 1.3 & 0.7 & 0.0 & 0.2 & 0.5 & 0.0 & 0.1 & 12.2 & 0.0 & 1.8 & 6.9 & 0.3 & 2.3 & 8.5 & 4.3 & 4.1 \\

Qwen3.5-35B-A3B      
& 4.1 & 1.8 & 1.2 & 0.5 & 0.0 & 0.3 & 0.2 & 0.0 & 0.1 & 11.4 & 0.0 & 1.1 & 4.6 & 0.0 & 1.2 & 3.5 & 1.2 & 1.3 \\

Gemma3-27B-IT       
& 0.3 & 0.3 & 0.1 & 0.7 & 0.0 & 0.2 & 0.0 & 0.0 & 0.0 & 3.3 & 0.0 & 0.3 & 3.1 & 0.0 & 0.7 & 3.9 & 1.7 & 1.8 \\

InternVL 3.5-38B    
& 1.0 & 0.5 & 0.2 & 0.2 & 0.2 & 0.2 & 0.2 & 0.0 & 0.1 & 1.9 & 0.0 & 0.3 & 0.3 & 0.0 & 0.1 & 2.1 & 1.7 & 1.1 \\

\hline
\multicolumn{19}{>{\columncolor{lightestgray}}c}{\textbf{SAGE}} \\
\hline
Claude Opus 4.6  
& 8.0 & 2.6 & 1.7 & 1.7 & 0.0 & 0.2 & 4.4 & 0.2 & 0.6 & 62.6 & 6.2 & 13.7 & 10.3 & 0.0 & 1.3 & 38.5 & 15.9 & 7.8 \\

Claude Sonnet 4.6  
& 2.3 & 1.0 & 0.5 & 0.2 & 0.0 & 0.1 & 0.0 & 0.0 & 0.0 & 26.0 & 5.7 & 7.7 & 13.9 & 0.5 & 2.8 & 5.6 & 2.7 & 1.5 \\

Kimi K2.5         
& 29.0 & 4.4 & 5.2 & 3.2 & 0.0 & 0.2 & 4.6 & 0.2 & 0.7 & 42.3 & 0.3 & 4.3 & 26.7 & 0.5 & 3.2 & 37.5 & 11.2 & 7.5 \\

Gemini 3 Pro       
& 18.5 & 9.0 & 7.6 & 3.0 & 1.0 & 1.7 & 7.8 & 2.2 & 3.0 & 6.5 & 0.5 & 2.1 & 8.0 & 1.8 & 3.9 & 17.8 & 6.6 & 9.3 \\

Llama4-Maverick-17B-IT       
& 0.0 & 0.0 & 0.0 & 0.2 & 0.0 & 0.1 & 0.0 & 0.0 & 0.0 & 0.0 & 0.0 & 0.0 & 0.0 & 0.0 & 0.0 & 0.0 & 0.0 & 0.0 \\

Qwen3.5-35B-A3B      
& 2.1 & 0.0 & 0.3 & 0.7 & 0.0 & 0.2 & 0.2 & 0.0 & 0.0 & 2.7 & 0.0 & 0.4 & 3.3 & 0.0 & 0.6 & 3.7 & 1.2 & 1.0 \\

Gemma3-27B-IT       
& 3.1 & 0.3 & 0.4 & 1.7 & 0.0 & 0.2 & 2.7 & 0.5 & 0.1 & 17.3 & 0.0 & 1.3 & 10.3 & 0.0 & 0.6 & 14.7 & 7.2 & 2.2 \\

InternVL 3.5-38B     
& 2.1 & 0.3 & 0.3 & 1.7 & 0.0 & 0.2 & 1.7 & 0.7 & 0.3 & 0.0 & 0.0 & 0.0 & 1.3 & 0.0 & 0.2 & 8.5 & 1.0 & 1.9 \\

\hline
\multicolumn{19}{>{\columncolor{lightestgray}}c}{\textbf{VERGE}} \\
\hline
Claude Opus 4.6 
& 28.3 & 3.9 & 5.0 & 1.2 & 0.0 & 0.2 & 4.1 & 0.0 & 0.8 & 73.4 & 14.1 & 16.4 & 10.0 & 0.3 & 1.2 & 34.0 & 12.4 & 7.6 \\

Claude Sonnet 4.6
& 12.3 & 3.9 & 3.1 & 0.5 & 0.0 & 0.2 & 0.7 & 0.0 & 0.1 & 65.6 & 10.3 & 17.3 & 24.4 & 0.5 & 4.8 & 27.3 & 8.5 & 7.5 \\

Kimi K2.5         
& 31.6 & 4.4 & 5.8 & 1.0 & 0.0 & 0.1 & 3.6 & 0.0 & 0.6 & 29.0 & 0.0 & 2.4 & 12.3 & 0.0 & 1.7 & 26.3 & 6.2 & 5.2 \\

Gemini 3 Pro       
& 4.6 & 1.5 & 1.6 & 4.0 & 1.0 & 2.2 & 2.7 & 0.7 & 1.2 & 1.9 & 0.3 & 0.6 & 14.4 & 5.9 & 7.1 & 17.0 & 7.7 & 9.2 \\

Llama4-Maverick-17B-IT       
& 2.6 & 0.5 & 0.6 & 1.0 & 0.0 & 0.4 & 0.0 & 0.0 & 0.0 & 10.0 & 0.3 & 1.3 & 4.4 & 0.3 & 1.0 & 9.1 & 3.1 & 3.0 \\

Qwen3.5-35B-A3B      
& 6.7 & 0.5 & 0.9 & 1.2 & 0.0 & 0.2 & 0.5 & 0.0 & 0.0 & 11.1 & 0.0 & 1.0 & 5.7 & 0.0 & 0.7 & 8.9 & 3.1 & 1.3 \\

Gemma3-27B-IT       
& 0.5 & 0.3 & 0.1 & 1.7 & 0.0 & 0.2 & 0.7 & 0.2 & 0.1 & 7.3 & 0.0 & 0.6 & 5.4 & 0.3 & 0.9 & 6.8 & 3.3 & 2.0 \\

InternVL 3.5-38B     
& 1.3 & 0.3 & 0.2 & 1.2 & 0.0 & 0.3 & 0.2 & 0.2 & 0.1 & 3.3 & 0.0 & 0.4 & 0.0 & 0.0 & 0.0 & 6.2 & 2.9 & 2.0 \\

\bottomrule
\end{tabular}
\label{tab:threshold_table_0.5}
\end{table*}


\end{document}